\documentclass[final,5p,times,twocolumn]{elsarticle}

\usepackage[utf8]{inputenc}
\usepackage[T1]{fontenc}

\usepackage{amssymb}

\usepackage{amsfonts}
\usepackage{amsmath}
\usepackage{algorithmic}
\usepackage{xcolor}
\usepackage{graphicx}

\usepackage{bbm}
\usepackage{tabularx}
\usepackage{adjustbox}
\usepackage{caption}

\usepackage{pgfplots}
\pgfplotsset{compat=1.18}	
\usetikzlibrary{pgfplots.statistics} 
\usetikzlibrary{matrix}
\usepgfplotslibrary{groupplots}

\usepackage[titletoc]{appendix}

\usepackage{tikz}
\usetikzlibrary{positioning, calc, chains}
\usetikzlibrary{decorations, decorations.pathreplacing, decorations.pathmorphing, calligraphy}
\usetikzlibrary{arrows.meta}
\usetikzlibrary{
    shapes,
    shapes.geometric,
    shapes.symbols,
    shapes.arrows,
    shapes.multipart,
    shapes.callouts,
    shapes.misc}
     
\tikzset{>=Stealth[round]}
\tikzset{
    ncbar angle/.initial=90,
    ncbar/.style={
        to path=(\tikztostart)
        -- ($(\tikztostart)!#1!\pgfkeysvalueof{/tikz/ncbar angle}:(\tikztotarget)$)
        -- ($(\tikztotarget)!($(\tikztostart)!#1!\pgfkeysvalueof{/tikz/ncbar angle}:(\tikztotarget)$)!\pgfkeysvalueof{/tikz/ncbar angle}:(\tikztostart)$)
        -- (\tikztotarget)
    },
    ncbar/.default=0.5cm,
}

\tikzset{square left brace/.style={ncbar=0.2cm}}
\tikzset{square right brace/.style={ncbar=-0.2cm}}

\usepackage{hyperref}
\usepackage{array}
\usepackage{multirow}
\usepackage{color, colortbl}
\usepackage{comment}
\usepackage{booktabs}
\usepackage{stfloats}

\usepackage{subfiles}
\usepackage{cleveref}
\usepackage{xr-hyper}

\usepackage{hhline}

\journal{AI Open}

\newcommand{\rr}{\color{black}}
\newcommand{\bb}{\color{black}}

\definecolor{Vgreen}{RGB}{141,211,199}
\definecolor{Vviolet}{RGB}{190,186,218}
\definecolor{Vred}{RGB}{251,128,114}
\definecolor{Vblue}{RGB}{128,177,211}
\definecolor{Vorange}{RGB}{253,180,98}
\definecolor{Vlime}{RGB}{179,222,105}
\definecolor{Vpink}{RGB}{252,205,229}
\definecolor{Vpurple}{RGB}{188,128,189}

\hypersetup{
  colorlinks=false,
  linkbordercolor=white,
  urlbordercolor=white,
  pdfborder={0 0 0}
}

\makeatletter
\renewcommand*\env@matrix[1][*\c@MaxMatrixCols c]{%
  \hskip -\arraycolsep
  \let\@ifnextchar\new@ifnextchar
  \array{#1}}

\makeatother

\newcolumntype{C}[1]{>{\hspace{0pt}\centering\arraybackslash}p{#1}}
\newcommand{\customemoji}[1]{\includegraphics[height=1em]{#1}}

\begin{document}






\begin{frontmatter}
    
\title{Not Another Imputation Method: A Transformer-based Model for Missing Values in Tabular Datasets}

\author[UCBM]{Camillo Maria Caruso} 
\ead{camillomaria.caruso@unicampus.it}

\author[UCBM,UMU]{Paolo Soda} 
\ead{p.soda@unicampus.it, paolo.soda@umu.se}

\author[UCBM]{Valerio Guarrasi\corref{cor1}} 
\ead{valerio.guarrasi@unicampus.it}
\cortext[cor1]{Corresponding author: valerio.guarrasi@unicampus.it}

  
\affiliation[UCBM]{organization={Research Unit of Computer Systems and Bioinformatics, Department of Engineering, Università Campus Bio-Medico di Roma},
            city={Roma},
            state={Italy},
            country={Europe}}

\affiliation[UMU]{organization={Department of Diagnostics and Intervention, Radiation Physics, Biomedical Engineering, Umeå University},
            city={Umeå},
            state={Sweden},
            country={Europe}}

\begin{abstract}

Handling missing values in tabular datasets presents a significant challenge in training and testing artificial intelligence models, an issue usually addressed using imputation techniques.
Here we introduce “Not Another Imputation Method” (NAIM), a novel transformer-based model specifically designed to address this issue without the need for traditional imputation techniques.
\rr NAIM's ability to avoid the necessity of imputing missing values and to effectively learn from available data relies on two main techniques: the use of feature-specific embeddings to encode both categorical and numerical features also handling missing inputs; the modification of the masked self-attention mechanism to completely mask out the contributions of missing data. \bb 
Additionally, a novel regularization technique is introduced to enhance the model’s generalization capability from incomplete data.
We extensively evaluated NAIM on 5 publicly available tabular datasets, demonstrating its superior performance over 6 state-of-the-art machine learning models and 5 deep learning models, each paired with 3 different imputation techniques when necessary. 
The results highlight the efficacy of NAIM in improving predictive performance and resilience in the presence of missing data.
To facilitate further research and practical application in handling missing data without traditional imputation methods, we made the code for NAIM available at \url{https://github.com/cosbidev/NAIM}.

\end{abstract}


\end{frontmatter}

\section{Introduction}\label{sec:introduction}

Tabular data, characterized by data structured in tables with rows and columns, i.e., samples and features, respectively, poses unique challenges in training and testing artificial intelligence~(AI) models because they differ from other data structures, such as text and speech that have a sequential nature or the images, which are characterized by spatial coherence.
\rr 
These challenges arise because, unlike in other domains, tabular data often consists of a \textit{heterogeneous mix of categorical and numerical features}. 
Additionally, each feature can have a \textit{different distribution and scale}, necessitating careful preprocessing steps such as normalization or embedding. 
Furthermore, tabular datasets frequently contain \textit{missing values}, requiring models that can robustly handle such inconsistencies.
\bb
To date, machine learning (ML) ruled over deep learning (DL) in many fields involving tabular data, offering methods capable of meeting each of these challenges~\cite{bib:review_ML, bib:review_XGB}. 

More specifically, the missing data issue poses a significant challenge in the realm of tabular data and common reasons for this event are: 
\textit{i}) human error, where inaccuracies during data entry or collection lead to gaps in information; \textit{ii}) non-response, common in surveys, where participants might skip questions for reasons like privacy or lack of interest; \textit{iii}) data corruption, caused by technical failures or errors; \textit{iv}) attrition, particularly in longitudinal studies, caused by participants drop out; \textit{v}) systematic loss, where data is missing under specific conditions.
Each of these factors could play a significant role in the challenge associated with handling missing values in tabular data: this absence of information may affect either training, testing, or both sets at the same time, whereas most of the state-of-the-art approaches need a complete dataset to function properly.
This brings the need for approaches that ignore the missing values and use all the available data to perform the task at hand.  
The state-of-the-art faced the problem using two main approaches~\cite{bib:review_ML, bib:review_XGB}: on the one hand, some methods fill in the missing entries, usually during the preprocessing phase before feeding the data to the model, by imputing the values using a predefined rule or an algorithm~\cite{bib:review_ML}; on the other hand, other approaches intrinsically handle missing data during inference time.

Recently, we have witnessed the successes of the transformer architecture in different domains. 
Since the transformer's core mechanism of self-attention has been adapted to computer vision~\cite{bib:VIT} and speech recognition~\cite{bib:whisper}, some studies have started adapting this architecture to tabular data~\cite{bib:tabnet, bib:tabtransformer, bib:fttransformer}; although, none has proposed specific solutions to handle missing values. 
This is why in this manuscript we present
\emph{``Not Another Imputation Method''} (NAIM), a novel transformer-based model for tabular data, specifically designed to face the missing values problem.
The main contributions of this work are:
\begin{itemize}
    \item The development of a transformer model that integrates feature-specific embeddings and a novel masked self-attention mechanism able to learn only from the available information without any need for imputation of missing values.
    \item The proposal of a novel regularization technique, which randomly masks each sample at every epoch. 
    This method, presenting a different sample version at each epoch, enables the model to generalize from incomplete data, ensuring more resilient and accurate predictive performance.
    \item A wide experimental assessment, showing that NAIM achieves better performance than state-of-the-art models in $5$ publicly available classification tasks. 
    State-of-the-art approaches consist of $6$ ML and \rr$5$ \bb DL models, paired with $3$ different imputation techniques, when necessary.
    Results indicate a noteworthy advancement of transformer and not transformer-based DL approaches over ML in addressing the domain of tabular data, highlighting its efficacy and potential in this area.
\end{itemize}

The manuscript is organized as follows: Section~\ref{sec:SOTA} presents the state-of-the-art of data imputation techniques and those models able to handle missing values (Section~\ref{sec:sota_missing}) and the main adaptations of transformers to the tabular data domain (Section~\ref{sec:transformer}). 
Section~\ref{sec:methods} introduces the proposed model, whereas Sections~\ref{sec:experimental_setup} and \ref{sec:results} discuss the experimental configuration and the results, respectively. 
Section~\ref{sec:conclusion} provides concluding remarks.


\section{State-of-the-art}\label{sec:SOTA}

Despite the advancements in handling missing values in tabular data, the state-of-the-art shows the following limitations. 
Current methods primarily focus on imputing missing entries or adapting models to work with incomplete data. 
However, these strategies often fail to fully leverage the inherent patterns in the data or require extensive preprocessing, potentially leading to information loss or biased predictions. 
Moreover, while the transformer architecture has shown promising results in various domains, its application to tabular data with missing values has not been realized, indicating a gap in effectively addressing this challenge within the deep learning domain.

This section first presents the most established strategies to address missing values in tabular data (Section~\ref{sec:sota_missing}) and then we explore existing transformer-based methods for tabular data (Section~\ref{sec:transformer}).
These methodologies are then employed as benchmarks in the experiments detailed in Sections~\ref{sec:experimental_setup} and \ref{sec:results}.

\subsection{SOTA techniques for missing values}\label{sec:sota_missing}

The simplest way to handle missing values is the \textit{Complete Case Analysis}~\cite{bib:complete_data} that completely excludes samples and/or features with missing values; despite its simplicity, its use can cause a considerable loss of information for the model, which cannot be tolerated to make AI models resilient to be used in practice.
\rr Consequently, several approaches have been developed to retain as much information as possible.

\subsubsection{ML approaches}\label{sec:sota_ML}
\bb
State-of-the-art techniques for managing missing values mainly focus either on imputing missing values~\cite{bib:review_ML, bib:complete_data, bib:trees} or on applying a specific strategy for missing entries~\cite{bib:review_ML, bib:trees}.
In~\cite{bib:review_ML}, where the authors benchmark different approaches to handle missing values, the imputation techniques are distinguished in \textit{Constant Imputation} and \textit{Conditional Imputation}. 
The first, despite its simplicity, can effectively support models by replacing missing values with a central tendency measure, e.g., mean imputation; the latter comprises techniques recognized for leveraging patterns within the data to predict missing values accurately, e.g., Multiple Imputation by Chained Equations (MICE)~\cite{bib:MICE} and K-Nearest Neighbors imputation (KNN)~\cite{bib:KNNimputer}. 
Furthermore, the authors in \cite{bib:review_ML} also introduce the \textit{Missing Incorporated in Attributes} (MIA) strategy, specifically advantageous for tree-based models, which incorporate missingness directly into the model, thus preserving all available data. 
By presenting some of the main possible approaches to handle missing values, this analysis underlines the multitude of available options and the need to identify the most suitable one for the specific task at hand. 
Furthermore, the prevalence of imputation approaches over models that can handle missing values, which are tree-based only, highlights the need for approaches that can ignore missing values rather than impute them.
Indeed, identifying the most suitable imputation method for a specific downstream task in advance is a considerable challenge.
Ideally, it would be better to rely solely on the data at hand, avoiding the introduction of potentially misleading information.
This approach, unlike imputation strategies that rely on the training set, would ensure a robust analysis, ignoring missing values entirely and making the most of the available data.

\rr 
\subsubsection{DL approaches}\label{sec:sota_DL}
DL approaches have introduced new perspectives and possibilities compared to traditional ML methods, particularly in handling complex and structured data scenarios. 
A representative example of this advancement in the missing data handling problem is GRAPE~\cite{bib:GRAPE}, a graph neural network (GNN)-based architecture specifically designed for tabular data analysis able to simultaneously perform feature imputation and label prediction. 
GRAPE, remapping the dataset onto a bipartite graph composed of sample nodes and feature nodes, intrinsically handles missing values by simply omitting edges between nodes where data is unavailable.
This setting lets GRAPE formulate feature imputation as an edge-level prediction problem and classification as a node-level prediction task.

Nevertheless, in recent years, transformer models, originally developed for natural language processing~(NLP)~\cite{bib:transformer}, have become predominant in various application domains, including computer vision~\cite{bib:VIT} and speech recognition~\cite{bib:whisper}. 
Although some extensions have been proposed to analyze tabular data, many transformer-based approaches still struggle to effectively manage missing data.
\bb

\subsection{Transformer for tabular data}\label{sec:transformer}

Currently, there are 3 main transformer-based approaches, each developing a different aspect of the transformer architecture to adapt it to tabular data: TabNet~\cite{bib:tabnet}, TabTransformer~\cite{bib:tabtransformer}, and FTTransformer~\cite{bib:fttransformer}. 

TabNet~\cite{bib:tabnet} leverages the self-attention mechanism to selectively focus on specific features for decision-making, which is crucial in interpreting high-dimensional data. 
Learning a sparse masking matrix and mimicking the boosting technique by adding a layer at each step, TabNet can perform a dynamic feature selection.
Its design facilitates interpretability, allowing it to offer insights into feature importance and decision pathways, a critical requirement in fields like finance and healthcare where understanding model decisions is as important as their accuracy.

TabTransformer~\cite{bib:tabtransformer} addresses the embedding of categorical features within tabular data. 
It applies a transformer-based self-attention mechanism to create contextual embeddings for categorical features, similar to how transformers process words in NLP. 
By doing so, it captures complex inter-feature relationships and dependencies, enhancing the model's predictive performance.
Given $x \in \mathbbm{R}^n$ as a mixture of categorical $x^{cat}$ and numerical $x^{num}$ features, all categorical features values $x^{cat}$ are encoded twice: on the one hand, a feature-specific lookup table $E_i^{pos}$ encodes the value with respect only to the possible categorical values of that feature; on the second hand all categorical values are encoded using a shared lookup table $E$ that embeds all the categorical features' values:
\begin{equation}
\scalebox{.9}{$
    e_i^{cat} = [E_i^{pos}(x_i^{cat}),E(x_i^{cat})].
$}
\end{equation}
Instead, the numerical features $x^{num}$ simply undergo a normalization step.
  
FTTransformer~\cite{bib:fttransformer} further explores the potential of transformer models in extracting patterns and interactions within tabular data, which are often overlooked by traditional machine learning methods. 
Specifically, this model addresses the embedding of numerical features applying two different approaches for the numerical and categorical features.
The embedding $e_i^{num}$ for a given numerical feature $x_i^{num}$ is computed as follows: 
\begin{equation}
\scalebox{.9}{$
\begin{aligned}
    e_i^{num} = b_i + x_i^{num}\cdot W_i^{num}, &&x_i^{num} \in \mathbbm{R}, &&e_i^{num} \in \mathbbm{R}^{d_e}
\end{aligned}
$}
\end{equation}
where the $d_e$-dimensional embedding vector is computed as the sum of the $i$-th feature bias, $b_i \in \mathbbm{R}^{d_e}$, and the multiplication of the feature value with the feature-specific vector $W_i^{num} \in \mathbbm{R}^{d_e}$.
Instead, the embedding $e_i^{cat}$ for a given categorical feature $x_i^{cat}$ is implemented as the lookup table $E_i^{cat}$, which associates a $d_e$-dimensional trainable vector to the $k_i$ entries of the specific categorical feature:
\begin{equation}
\scalebox{.9}{$
\begin{aligned}
     e_i^{cat} = b_i + E_i^{cat}(x_i^{cat}), &&x_i^{cat} \in \mathbbm{N}_{k_i}, &&e_i^{cat} \in \mathbbm{R}^{d_e}.
\end{aligned}
$}
\end{equation}

\rr 
\subsection{Current limitations}
As illustrated in the previous sections, current state-of-the-art methods still have several limitations. 
Among traditional ML approaches, none of the methods truly ignore missing values; the MIA strategy, which does not impute missing data, attempts to incorporate missingness into the learning process rather than bypassing it completely. 
Regarding DL, GRAPE is the only architecture that can explicitly ignore missing features, but its dependence on GNN structures limits its integration and scalability. 
In contrast, transformer-based models, not having dedicated embeddings and a masking mechanism, cannot handle missing data. 
This shortcoming highlights the need for transformer-based solutions that can robustly ignore missing features while maintaining architectural flexibility and compatibility with broader modeling frameworks.
\bb

\section{Methods}\label{sec:methods}

This section first \rr presents a formal definition of the problem; then it \bb examines the proposed NAIM architecture (Section~\ref{sec:NAIM}), presenting the innovation we made to the embedding of the features and the novel masked self-attention mechanism to dynamically handle missing values; finally, in section \ref{sec:augmentation}, we introduce a novel regularization strategy that aims to enable the model to learn how to handle missing data even if they are not present in the training set, and to extract the optimal representation for each feature by preventing any co-adaptation between them. 

\rr 
\subsection{Problem definition}
Let $X \in \mathbb{R}^{n \times m}$ represent a data matrix with $n$ samples and $m$ features, where the element $x_{i}^{num}$ corresponds to a numerical feature of the $i$-th sample, while $x_{i}^{cat}$ represents a categorical feature of the same sample. 
In scenarios involving missing data, some of these feature values are unavailable. 
This is captured, for each sample separately, by a binary mask matrix $M \in \{-\infty,0\}^{m \times m}$, where a value of $M_{ij} = 0$ indicates that $x_{ij}$ is observed.
Typically, datasets are also associated with labels for a specific downstream task, which we denote as $Y \in \mathbb{R}^n$. 
Our focus is on predicting labels for test instances. 
\bb

\subsection{NAIM architecture}\label{sec:NAIM}

\begin{figure*}[!ht]
    \centering
    \includegraphics[width=\textwidth]{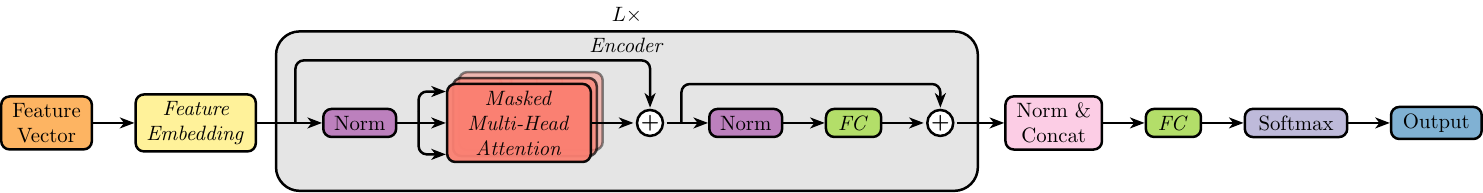}
    \caption{The architecture of NAIM, composed of the \textit{Feature Embedding}, the $\mathit{Encoder}$ equipped with the \textit{Masked Multi-Head Attention} mechanism, and the final classification head~\rr\cite{bib:transformer}\bb.}
    \label{fig:model}
\end{figure*}

The transformer architecture~\cite{bib:transformer}, originally developed for text translation, includes two main blocks: the first, known as the $\mathit{encoder}$, extracts a hidden representation from the data, whilst the second, named the $\mathit{decoder}$, reconstructs the hidden representation in the context of the specific domain. 
For instance, in the NLP domain, the decoder generates the next token or performs complex interactions given an input sequence. 
Nevertheless, the decoder may not be required in scenarios where decision-making is the primary objective.
Given that most of the tasks related to tabular data are classification or regression, the transformer-based models mostly make use of the $\mathit{encoder}$ part only with a fully-connected ($\mathit{FC}$) module for the final prediction~\cite{bib:tabnet, bib:tabtransformer, bib:fttransformer}.
However, to perform other tasks, e.g., features imputation or self-supervised pretraining, the $\mathit{decoder}$ part could also be used to reconstruct the feature representation in another domain~\cite{bib:tabnet}.
Indeed, when developing NAIM to get a model that handles missing features and learns from the available information without any need for imputation to perform a classification task, we opted for the encoder-only architecture followed by a $\mathit{FC}$ module (\figurename~\ref{fig:model}), which receives as input the normalized and concatenated version of the $\mathit{encoder}$ output (grey block).

Our approach exploits the \textit{Feature Embedding} (yellow block) and \textit{Masked Multi-Head Attention} (red block) steps reported in the figure.
More specifically, our idea stems from observing that the use of the padding index in the lookup table paired with the masked self-attention mechanism, could be an interesting and unexplored way to handle missing entries for both the categorical and numerical types of features in tabular data.

\begin{figure}[!ht]
    \centering
    \includegraphics[width=\columnwidth]{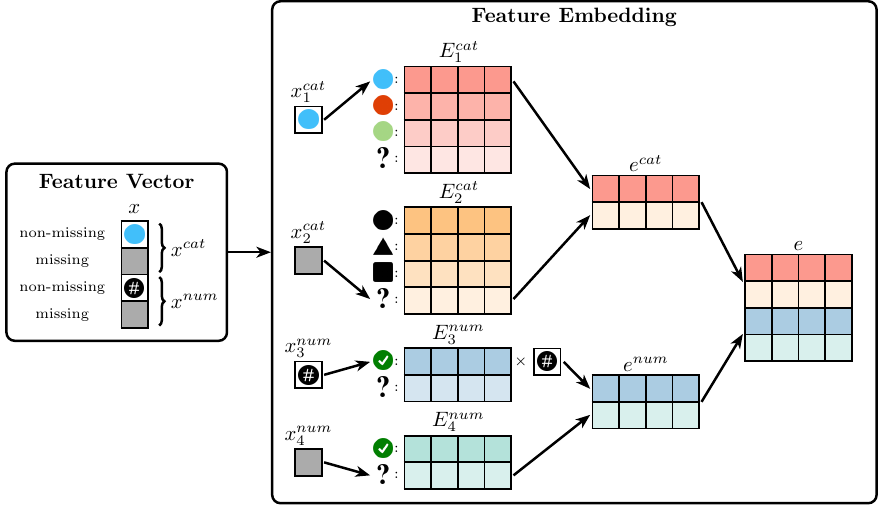}
    \caption{The proposed \textit{Feature Embedding} process for tabular data. In the example, the feature vector $x$ has $4$ features: $2$ categorical ($x^{cat}$) and $2$ numerical ($x^{num}$). The colors (\customemoji{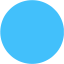}, \customemoji{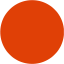}, \customemoji{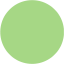}) and the shapes (\customemoji{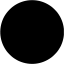}, \customemoji{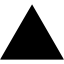}, \customemoji{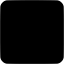})
    are examples of possible values for the first categorical feature $x_1^{cat}$ and for the second one $x_2^{cat}$, respectively, whilst \customemoji{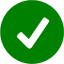} stands for a non-missing numerical feature and \customemoji{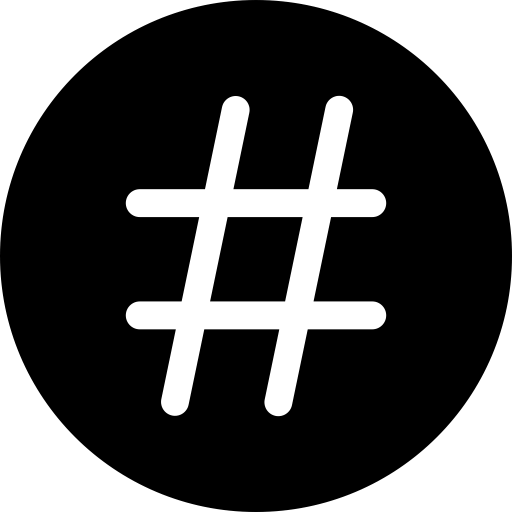} for its value. Finally, \customemoji{missing} indicates the padding index related to missing features for both types of features.
    In the \textit{Feature Embedding} block, we can see how the embedding $e$ of the feature vector $x$ is composed of the concatenation of embedded representations of the categorical and numerical features, denoted as $e^{cat}$ and $e^{num}$, respectively.
    These representations are composed by the concatenation of the vectors associated with each feature value, selected using the feature-specific lookup tables $E_i^{cat}$ and $E_i^{num}$.}
    \label{fig:embedding}
\end{figure}

Focusing on the \textit{Feature Embedding}, which is a fundamental step to find a richer representation of the input data useful in the downstream task, we observed that the use of the padding index is reported in literature~\cite{bib:tabtransformer, bib:fttransformer}, but none are able to handle missing values for both categorical and numerical features.
Indeed, with the embedding lookup table $E_i$ paired with the masked self-attention, we can completely exclude the missing values from the gradient calculations defining the $k_i+1^{th}$ entry, named \textit{padding index} or \guilsinglleft pad\guilsinglright, which is assigned to a not-trainable vector of zeros.
Using the notation introduced in section~\ref{sec:transformer}, the embedding of categorical features $e_i^{cat}$ is given by:
\begin{equation}\label{eq:cat_emb}
\scalebox{.9}{$
\begin{aligned}
     e_i^{cat} = b_i + E_i^{cat}(x_i^{cat}), &&x_i^{cat} \in \mathbbm{N}_{k_i}, &&e_i^{cat} \in \mathbbm{R}^{d_e}
\end{aligned}
$}
\end{equation} \bb
whereas we defined an embedding lookup table $E_i^{num}$ for each numerical feature with 2 possible entries, named \textit{present} and \textit{missing}.
Using the padding index of the lookup table, these entries are associated with a trainable and a not-trainable $d_e$-dimensional vector, respectively. 
Then, we scaled the numerical embedding vector $e_i^{num}$ by multiplying it by the feature value $x_i^{num}$ when present:
\begin{equation}\label{eq:num_emb}
\scalebox{.8}{$
\begin{aligned}
    e_i^{num} = b_i + x_i^{num} \cdot E_i^{num}(x_i^{num}), && x_i^{num} \in \mathbbm{R}, && e_i^{num} \in \mathbbm{R}^{d_e}.
\end{aligned}$}
\end{equation}
This setup, further described with an example reported in \figurename~\ref{fig:embedding}, ensures that both types of features can be handled when presenting missing values without impacting the learning process.
The figure provides an example of a feature vector with 2 categorical ($x^{cat}$) and 2 numerical ($x^{num}$) features. 
For each of these 2 types of features, there are one missing and one non-missing value, which are then embedded using the respective look-up tables in the \textit{Feature Embedding} block. 
In particular, we can notice that both the missing features are encoded using the entry relative to the missing value of the look-up tables, represented with the \customemoji{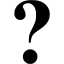} symbol.
After obtaining the embedding vectors of the features, these are concatenated in the embedded representation $e$, which will be fed to the encoder later in the model architecture.

Focusing on the masked self-attention mechanism to adapt it effectively for tabular data and to completely mask out the contribution of the missing features, we first argued that it is necessary to modify it.
Indeed, masked self-attention, a variant of the standard attention mechanism, plays a key role: it allows the model to focus only on certain parts of the input sequence while ignoring others. 
Traditionally, after the \textit{Feature Embedding} step, the transformer calculates the query, key, and value matrices, denoted as $Q$, $K$, and $V$ respectively, through linear transformations of $e$, i.e., mapping the embedding into a smaller space $d_h$ based on the chosen number of heads $h$: 
\begin{equation}\label{eq:embeddings}
\scalebox{.9}{$
    \left\{
    \begin{aligned}
    Q &= e \cdot W^Q, &&&d_h=d_e/h \\
    K &= e \cdot W^K, &&&W^Q, W^K, W^V \in \mathbbm{R}^{d_e \times d_h} \\
    V &= e \cdot W^V, &&&Q, K, V \in \mathbbm{R}^{n \times d_h} 
    \end{aligned}
    \right.
$}
\end{equation}
where $W^Q$, $W^K$ and $W^V$ are weights matrices learned during training. 
Then, in the standard masked self-attention, 
\begin{equation}\label{eq:masked_attn}
\scalebox{.9}{$
\mathit{Attention}(Q,K,V) = \underbrace{\mathit{softmax} \left(\frac{QK^T}{\sqrt{d_h}} + M\right)}_{A \in \mathbbm{R}^{n \times n}}V $}
\end{equation}
the attention matrix $A$ uses the masking matrix $M \in \mathbbm{R}^{n\times n}$ that sums $-\infty$ to the weights that should be ignored, zeroing their influence after the $\mathit{softmax}$ operation. 
In the NLP domain, this technique is mainly employed to achieve $2$ different purposes: on the one hand, the use of a mask $M$ such as
\begin{equation}
\scalebox{.73}{$
\begin{aligned}
    M_{ij} = \left\{  
    \begin{aligned}
        &0 &\text{if}~i \leq j\\
        -&\infty &\text{if}~i > j
    \end{aligned}
    \right.,
    &&M = 
    \renewcommand\arraystretch{1.2}\begin{bmatrix}[*5{C{5.6mm}}]
                0 &     -$\infty$ &     ... &   -$\infty$ &     -$\infty$ \\
                0 &             0 &     ... &   -$\infty$ &     -$\infty$ \\
              \rotatebox{90}{...} &           \rotatebox{90}{...} &     \large\rotatebox{135}{...} &         \rotatebox{90}{...} &           \rotatebox{90}{...} \\
                0 &             0 &     ... &           0 &     -$\infty$ \\
                0 &             0 &     ... &           0 &             0 \\
                \end{bmatrix}
\end{aligned}$}
\end{equation}
allows the model to ignore future positions in the sequence, avoiding any leakage of information when training the model with the same sentence at different stages of completion; on the other hand, it could be used a mask $M$ that cancels out the contributions of the last tokens
\begin{equation}\label{eq:pad}
\scalebox{.73}{$
\begin{aligned}
    M_{ij} = \left\{  
    \begin{aligned}
        &0 &\text{if}~x_j\neq\text{\guilsinglleft pad\guilsinglright} \\
        -&\infty &\text{if}~x_j=\text{\guilsinglleft pad\guilsinglright}
    \end{aligned}
    \right.,
    \quad
    &M = 
    \renewcommand\arraystretch{1.2}\begin{bmatrix}[*5{C{5.6mm}}]
                0 &             0 &     ... &           0 &     -$\infty$ \\
                0 &             0 &     ... &           0 &     -$\infty$ \\
              \rotatebox{90}{...} &           \rotatebox{90}{...} &     \large\rotatebox{135}{...} &         \rotatebox{90}{...} &           \rotatebox{90}{...} \\
                0 &             0 &     ... &           0 &     -$\infty$ \\
                0 &             0 &     ... &           0 &     -$\infty$ \\
                \end{bmatrix}
\end{aligned}$}
\end{equation}
allowing the model to analyze sequences shorter than the dimension $n$ used for training.

\begin{figure}[!ht]
    \centering
    \includegraphics[width=\columnwidth]{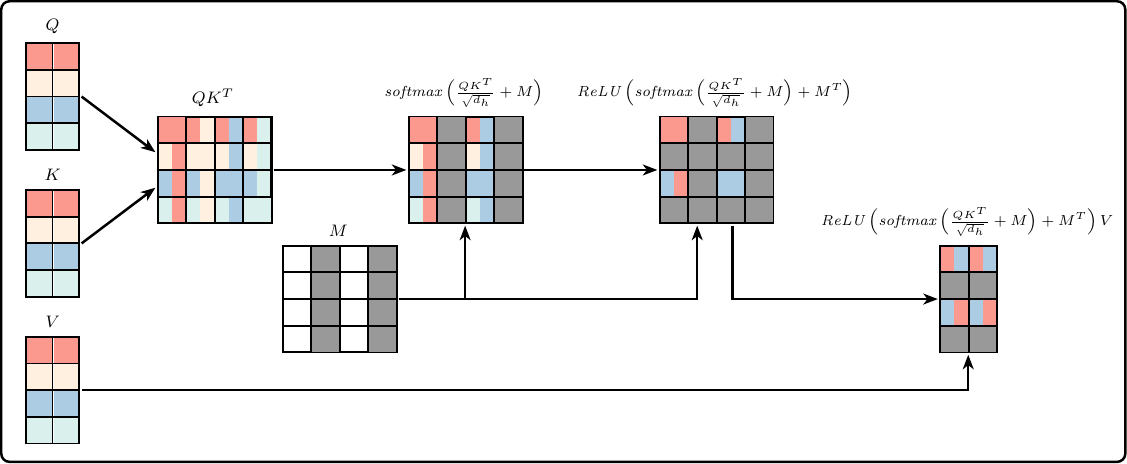}
    \caption{The proposed masked self-attention mechanism, designed to effectively ignore the impact of missing entries within the attention matrix. In the example, the $QK^T$ matrix, obtained by the multiplication of the $Q$ and $K$ representations, is reported as an example of how the contributions of the different features, identified by different colors, mix up together.  
    Next, the classic masked self-attention mechanism is applied, and some of the contributions of the missing features (indicated with \customemoji{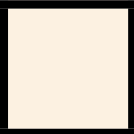} and \customemoji{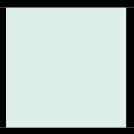}) remain. 
    Then, the proposed attention mechanism ensures that the influence of these missing values is completely masked out, before the multiplication by the representation $V$ of the sample.}
    \label{fig:mask}
\end{figure}

Considering the structure of these 2 possible masks, it is straightforward that only the latter could be adapted to mask missing features, since the former would mask large portions of the samples, even related to features that are non-missing. 
On this ground, the use of the Eq.~\ref{eq:masked_attn} paired with the mask $M$ in  Eq.~\ref{eq:pad} used to mask the columns related to the missing values, might seem a viable solution, but, as shown in \figurename~\ref{fig:mask} with colors and reported in~\ref{app:mask} with an example, it does not cancel out the undesired value.
Indeed, in \figurename~\ref{fig:mask} we observe both the application of traditional masked attention and our new proposal: starting from left to right we have the matrices $Q$, $K$ and $V$ generated from the example feature vector shown in \figurename~\ref{fig:embedding}, where the second and fourth features were missing.
Immediately afterward, we report the attention matrix resulting from the calculation of the $QK^T$ product, which is useful for understanding the next steps and especially how the contributions of the different features mix up together. 
In the next step, we show that the traditional masked attention mechanism, by masking the second and fourth columns, eliminates some of the contributions of the missing features, but leaves others on the second and fourth rows. 
For this reason, we here propose a new masked self-attention designed to completely mask out the contributions of the missing values using twice the same mask $M$ from equation~\ref{eq:pad}:
\begin{equation}\label{eq:new_attn}
\scalebox{.8}{$
    \mathit{Attention}(Q,K,V) = \mathit{ReLU} \left( \mathit{softmax} \left(\frac{QK^T}{\sqrt{d_h}} + M\right) + M^T \right)V $}
\end{equation}
In this way, the attention matrix's rows and columns related to missing values are now assigned zero attention, ignoring their contribution. 

\subsection{Regularization technique}\label{sec:augmentation}

\begin{figure}[!ht]
    \centering
    \includegraphics[width=.85\columnwidth]{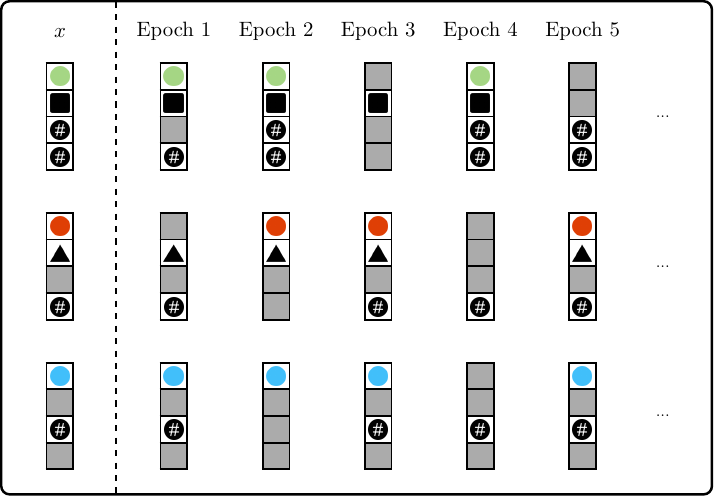}
    \caption{The proposed regularization strategy performed at every epoch before feeding the sample to the model. 
    The colors (\customemoji{blue}, \customemoji{red}, \customemoji{green}) and the shapes (\customemoji{triangle}, \customemoji{square})
    are examples of possible values for the categorical features, whilst \customemoji{number} stands for a numerical feature.
    In the example are reported $3$ feature vectors and their masked versions created in $5$ different epochs. It should be noted that, when some features are originally missing, only the non-missing entries can be masked.}
    \label{fig:augmentation}
\end{figure}

Given the capacity of NAIM to effectively handle missing features, we introduce a novel regularization technique designed to enhance model robustness by simulating missing data scenarios. 
This approach, inspired by the Cutout technique~\cite{bib:cutout} randomly masks elements within feature vectors (\figurename~\ref{fig:augmentation}), taking into account real-world data's inherent variability and incompleteness.
By integrating this regularization technique, we aim to improve the model's ability to generalize from incomplete patterns, thus enhancing predictive performance on datasets with varying degrees of missing information.

Given the feature vector $x \in \mathbb{R}^n$, where $n$ represents the dimensionality of the vector, let $v \leq n$ be the number of non-missing elements. 
The vector may be fully populated with $v=n$ or contain missing values with $v<n$. 
The process is governed by a binary decision variable $B \sim \text{Bernoulli}(0.5)$, which determines whether the masking will be applied to the instance $x$. 
If a sample is selected for masking, a random count of $c$ elements to mask is chosen uniformly from the set $\{1, 2, \ldots, v-1\}$, ensuring that at least one element remains unmasked.
Finally, $c$ non-missing elements within $x$ are randomly chosen and set their values to \textit{missing}, resulting in the augmented vector.

The masking operation simulates scenarios of incomplete data, a common challenge in practical applications. 
This methodological approach is systematically applied during the training phase, ensuring that each instance $x$ underwent the described random masking process with a probability of $50\%$: this introduces variability in the input data, encouraging the model to learn more robust features that are not overly dependent on the presence of specific elements within the feature vector.

\section{Experimental Configuration}\label{sec:experimental_setup}

In this section, we first report the data used in the experiments, the preprocessing applied and the metric used in the evaluation (Section~\ref{sec:data}), then we list the combinations of models and imputers used as competitors (Section~\ref{sec:competitors}).
We provide some implementation details about the parameters used during training in~\ref{app:training}. 

\subsection{Data and Evaluation}\label{sec:data}
We evaluate NAIM and competitor models on $5$ publicly available datasets (reported in \tablename~\ref{tab:data}) from the UCI repository~\cite{bib:UCI}.
We selected these datasets as they are used to benchmark models for tabular data in~\cite{bib:tabtransformer}:

\begin{itemize}
    \item \textbf{Adult}~\cite{bib:adult}: it is a dataset focusing on income prediction, which tries to predict people that make over $50k$ a year, from those that earn less than that amount, based on personal attributes such as education, occupation, and hours per week worked;
    \item \textbf{BankMarketing}~\cite{bib:bankmarketing}: its task is to predict whether the client will subscribe to a term deposit or not, and it comprises data from direct marketing campaigns of a Portuguese banking institution;
    \item \textbf{OnlineShoppers}~\cite{bib:onlineshoppers}: it involves predicting whether a visitor will make a purchase based on session information, like page views and time spent on the site;
    \item \textbf{SeismicBumps}~\cite{bib:seismicbumps}: it aims at predicting seismic hazards in coal mines, i.e., whether any seismic bump with an energy higher than $10^4~J$ was registered in the next $8$ hours, based on seismic activity and energy measurements;
    \item \textbf{Spambase}~\cite{bib:spambase}: it involves identifying whether an email is spam or not based on word frequencies and other email attributes.
\end{itemize} 
We note that this selection of datasets shares the same type of task, i.e., classification, and that all of them are complete, except for Adult, which has only $1\%$ of missing data.
The datasets exemplify well-curated data; however, in practical real-world scenarios, this level of completeness may not always be achievable. 

For each dataset, we normalize the numerical features in the range $[0, 1]$ and apply one-hot encoding to the categorical features, before feeding them to the models.
In the experiments featuring NAIM, HistGradientBoost, TabNet, TabTransformer, FTTransformer and XGBoost, we do not apply the one-hot encoding, since their implementations can handle categorical features.
The preprocessing step is calibrated on the training data and then applied to validation and testing sets.

\begin{table*}[t]
    \centering
    \resizebox{.8\textwidth}{!}{
    \begin{tabular}{c|c|c|c|ll}
         \toprule
         \textbf{Dataset} & \textbf{\# of samples} & \textbf{\# of features} & \textbf{\# of categorical features} & \multicolumn{2}{c}{\textbf{Class distribution}} \\ 
         \midrule
         Adult \cite{bib:adult}                    & 48842 & 14 & 8 & $\mathbf{0}$: $37155$ & $\mathbf{1}$: $11687$ \\ 
         BankMarketing \cite{bib:bankmarketing}    & 41188 & 20 & 10 & $\mathbf{0}$: $36548$ & $\mathbf{1}$: $4640$ \\ 
         OnlineShoppers \cite{bib:onlineshoppers}  & 12330 & 17 & 7 & $\mathbf{0}$: $10422$ & $\mathbf{1}$: $1908$ \\
         SeismicBumps \cite{bib:seismicbumps}      &  2584 & 18 & 4 & $\mathbf{0}$: $2414$ & $\mathbf{1}$: $170$ \\ 
         Spambase \cite{bib:spambase}              &  4601 & 57 & 0 & $\mathbf{0}$: $2788$ & $\mathbf{1}$: $1813$ \\ 
         \bottomrule
    \end{tabular}}
    \caption{Datasets' details and references. Datasets' information consists of the number of samples, the number of features, how many of these features are categorical, and the classes' distribution.}
    \label{tab:data}
\end{table*}

Our experiments focus on artificially generated \emph{Missing Completely At Random} (MCAR) values, \rr the most common evaluation regime used in missing
data papers~\cite{bib:GRAPE}. \bb 
Indeed, MCAR's lack of systematic bias means that the remaining data can be considered a fair representation of the whole.
We aim to test our models under various missing data scenarios by introducing missing values at different percentages (denoted as $p$) across both training and testing sets. 
More specifically, we artificially generated missing percentages in the training and test sets separately, equal to $0\%$, $5\%$, $10\%$, $25\%$, $50\%$ and $75\%$ of the data in the set, for a total of $36$ possible variations with repetitions. 
This allows us to methodically delineate varying scenarios of data completeness and missingness: \textit{i}) a scenario with a well-curated data collection for both training and testing sets, i.e., $0\%$ of missing data in training and in testing; \textit{ii}) a well-curated collection for the training set followed by a testing set with missing values, e.g., $0\%$ of missing values in training and $25\%$ in testing; \textit{iii}), a scenario characterized by faulty data collection practices leading to missing values in the training phase, yet a completely curated test set without missing values, e.g., more common cases with $5\%$, $10\%$, or $25\%$ of missing data in training and $0\%$ in testing, or even more extreme cases where $50\%$ or $75\%$ of data are missing in the training set, but in the test set they are all not missing; \textit{iv}) scenarios, which could represent domains where complete data collection is not possible, characterized by missing values in both training and testing sets, e.g., situations in which low percentages such as $5\%$, $10\%$ or $25\%$ of the data are missing in both sets, or even more extreme circumstances in which $50\%$ or $75\%$ of the values are missing.
Given a targeted missing data percentage $p$ and the total number of samples $j$ in the considered set, we calculated the total number of values to be masked ($j \cdot n \cdot p$), considering any pre-existing missing values ($j\cdot n \cdot p - \sum_j (n - v_j)$).
Following this, a random masking matrix of dimension $j \times n$ is generated to match the set structure. 
We ensured that at least one value in any fully masked row or column was replaced, avoiding complete data loss in any specific dimension. 
This approach leads to samples and features exhibiting varied percentages of missing values, all adhering to the MCAR paradigm.

\subsection{Competitors}\label{sec:competitors}

We conduct an extensive comparison of our methodology against approaches that incorporate missing data imputation as preprocessing followed by model training, and against those models able to intrinsically handle missing values. 
In this respect, our analysis exploits $11$ unique model competitors, each integrated with the $3$ main imputation techniques, i.e., 
the mean constant imputation, the KNN and the MICE with their default parameters~\cite{bib:review_ML}.
This is summarized in \tablename~\ref{tab:competitors}, where the first column distinguishes between ML and DL approaches, the second lists the base learners, columns $3-5$ report the imputation techniques, and finally last column shows the use of an intrinsic strategy. 
In this table, each of the $35$ competitors is therefore marked by an ``$\times$''.
\begin{table}[!ht]
    \centering
    \resizebox{.9\columnwidth}{!}{
    \setlength\extrarowheight{1.5pt}\begin{tabular}{c|c|c|c|c|c}
        \toprule
        & & \multicolumn{3}{c|}{\textbf{Imputers}} & \\
         \textbf{Type} & \textbf{Model} & Mean Constant & KNN & MICE & Intrinsic \\
         \midrule
         \multirow{6}{*}{\rotatebox[origin=c]{90}{\makebox[0pt][c]{\parbox{2cm}{\centering\textbf{Machine \\Learning}}}}} & AdaBoost~\cite{bib:adaboost} & \large{$\times$} & \large{$\times$} & \large{$\times$} & \\\cline{3-6}
         & Decision Tree~\cite{bib:DT} & \large{$\times$} & \large{$\times$} & \large{$\times$} & \large{$\times$} \\\cline{3-6}
         & HistGradientBoost~\cite{bib:review_ML} & \large{$\times$} & \large{$\times$} & \large{$\times$} & \large{$\times$} \\\cline{3-6}
         & Random Forest~\cite{bib:RF} & \large{$\times$} & \large{$\times$} & \large{$\times$} & \large{$\times$} \\\cline{3-6}
         & SVM~\cite{bib:SVM} & \large{$\times$} & \large{$\times$} & \large{$\times$} & \\\cline{3-6}
         & XGBoost~\cite{bib:xgboost} & \large{$\times$} & \large{$\times$} & \large{$\times$} & \large{$\times$} \\
         \midrule
         \multirow{4}{*}{\rotatebox[origin=c]{90}{\makebox[0pt][c]{\parbox{2cm}{\centering\textbf{Deep \\Learning}}}}} & MLP~\cite{bib:mlp} & \large{$\times$} & \large{$\times$} & \large{$\times$} & \\\cline{3-6}
         & GRAPE~\cite{bib:GRAPE} &   &   &  & \large{$\times$} \\\cline{3-6}
         & TabNet~\cite{bib:tabnet} & \large{$\times$} & \large{$\times$} & \large{$\times$} & \\\cline{3-6}
         & TabTransformer~\cite{bib:tabtransformer} & \large{$\times$} & \large{$\times$} & \large{$\times$} & \\\cline{3-6}
         & FTTransformer~\cite{bib:fttransformer} & \large{$\times$} & \large{$\times$} & \large{$\times$} & \\
         \bottomrule
    \end{tabular}}
    \caption{Combinations of models and imputers used as competitors in the experiments. Each competitor is represented by an ``$\times$'', given by the combination of a learner and an imputation technique, further to intrinsic strategies.} 
    \label{tab:competitors}
\end{table}

We can categorize the competing models based on the subsequent analyses we intend to conduct. 
Our objective is to benchmark NAIM against leading methodologies in the field of missing tabular data, specifically machine learning models combined with imputation techniques. 
Listed alphabetically, these models include:

\begin{itemize}
\item \textbf{AdaBoost}~\cite{bib:adaboost}: it is employed as a cascade of classifiers, which is particularly effective for its ability to adaptively focus on hard-to-classify instances, enhancing overall model performance. 
This characteristic makes it a robust choice for handling diverse datasets, including those with imputed values.

\item \textbf{Decision Tree}~\cite{bib:DT}: it is a versatile machine learning algorithm particularly useful for analyzing tabular datasets. 
It excels in its interpretability, as it visually represents decision-making processes and variable importance, enabling straightforward insights into complex data relationships and patterns.

\item \textbf{HistGradientBoost}~\cite{bib:review_ML}: a variation of a gradient-boosted tree model, recognized for its efficiency in handling large datasets and its ability to improve upon the limitations of traditional gradient boosting by optimizing for speed and memory usage. 
Its inclusion allows for the assessment of advanced boosting techniques in missing data scenarios.

\item \textbf{Random Forest}~\cite{bib:RF}: an ensemble of decision trees, it is noted for its robustness against overfitting. 
Leveraging a multitude of decision trees enhances the model's accuracy and reliability, making it an exemplary model for evaluating the strengths of ensemble methods in the tabular data domain.

\item \textbf{Support Vector Machine} (SVM)~\cite{bib:SVM}: it is included for its versatility in dealing with non-linear data separations, offering, as a kernel machine equipped with the RBF kernel, a contrast to tree-based models in handling imputed datasets. 
The SVM’s capacity to project data into higher dimensions where classes are more easily separable makes it a valuable model for comparison.

\item \textbf{XGBoost}~\cite{bib:xgboost}: a sophisticated variation of AdaBoost that employs a gradient descent procedure to minimize the loss when adding weak learners, it stands out for its exceptional performance with tabular data. 
Its advanced handling of regularization and scalability positions XGBoost as the leading model for comparison in studies involving tabular datasets~\cite{bib:review_XGB}.
\end{itemize}
This selection of models, each with unique approaches for handling tabular data, provides a comprehensive background for evaluating the performance of our proposed methodology.
Furthermore, considering that Decision Tree, HistGradientBoost, Random Forest and XGBoost offer implementations incorporating the MIA strategy, we assessed their effectiveness in managing missing values.

Lastly, we aim to showcase the performance of our method in comparison to other DL models specifically designed for tabular data.
Our comparative analysis extends to $5$ advanced architectures, each selected for its novel approach to handling tabular data, particularly when combined with imputation techniques for managing missing values. 
These models, discussed in sections~\ref{sec:sota_DL} and \ref{sec:transformer}, bring unique perspectives and methodologies to the challenges of tabular data analysis:
\begin{itemize}
\item \textbf{Multilayer Perceptron} (MLP)~\cite{bib:mlp}: a foundational DL model, it stands out for its simplicity and versatility. 
It consists of multiple layers of neurons, each fully connected to those in the next layer, enabling the model to capture complex nonlinear relationships between features. 
Despite its straightforward structure, MLP's performance hinges on the quality of input data, making it crucial to pair with effective imputation techniques to address missing values.

\rr
\item \textbf{GRAPE}~\cite{bib:GRAPE}: \rr designed explicitly to tackle missing data within tabular datasets, this model employs a graph-based approach, representing each data instance as a node and modeling feature interactions as edges. 
Leveraging GNNs, it propagates information effectively across the nodes, allowing the model to learn directly from incomplete data without preliminary imputation. 
This structure enables GRAPE to capture complex relationships inherent to tabular data, enhancing robustness and predictive performance even in the presence of extensive missingness. 
\bb

\item \textbf{TabNet}~\cite{bib:tabnet}: leveraging the self-attention mechanism, it dynamically selects which features to focus on for making decisions. 
Its ability to perform dynamic feature selection mimics the boosting technique, enhancing its interpretability.
This model is particularly suited to interpreting high-dimensional data, offering insights into feature importance and decision pathways. 

\item \textbf{TabTransformer}~\cite{bib:tabtransformer}: this model innovates by embedding categorical features within tabular data, applying transformer-based self-attention mechanisms to create contextual embeddings. 
It captures complex inter-feature relationships and dependencies, significantly improving the predictive performance on tabular datasets. 

\item \textbf{FTTransformer}~\cite{bib:fttransformer}: it further explores the potential of transformers in extracting patterns and interactions within tabular data, employing distinct embedding strategies for numerical and categorical features.
\end{itemize}
These DL models, chosen for their innovative handling of tabular data with respect to classical ML approaches, are integrated with $3$ main imputation techniques since none of them can handle missing values for both categorical and numerical features. 

\subsection{Evaluation Metrics and Statistical Analysis}\label{sec:metrics_stats}

For evaluation, each dataset is divided into $5$ stratified cross-validation splits, to maintain the original class distribution, and for each fold, $20\%$ of the training samples are selected for validation.
We evaluate each experiment averaging out the values of Area Under the ROC curve (AUC) computed in the different cross-validation folds.
AUC is a valuable metric to evaluate classification tasks, since it represents the degree to which a model can correctly classify positive and negative instances across all possible thresholds, making it a comprehensive measure of model performance, even in case of imbalanced data~\cite{bib:AUC}.

To ensure the statistical validity of our results, we investigated the statistical differences between the predictions made by NAIM and those by the competitors using the Wilcoxon signed-rank test. 
To this end, we assessed the rate of experiments for each dataset individually, among the $36$ variations with repetitions of the $6$ percentages of missing values artificially generated, in which NAIM achieved statistically superior and inferior results compared to the competitors, setting $p = 0.05$.


\section{Results and Discussions}\label{sec:results}

As outlined in the previous section, we compare NAIM with $35$ leading competitors in ML and DL for tabular data, and we test the performance with $36$ different combinations of percentages of missing values both in the training and testing sets, across the various datasets, resulting in a total of $6480$ experiments, $1296$ per dataset. 
Therefore, in analyzing the performance of the different approaches, we considered each level of missing data within the training set separately, delineating performance metrics as the percentages of missing data in the testing set increased. 
In \tablename~\ref{tab:TABELLONE_RISULTATI} we reported for each combination of model and strategy to handle missing values (reported in different rows), the average performance, in terms of AUC and standard error, across the $5$ datasets under consideration, obtained at the different percentages of missing values (reported in the different columns). 
In particular, the first rows denote the percentage of missing values used in training and the specific percentage of missing data in the test set.
Similarly, in~\ref{app:results}, we also reported the detailed tables per dataset (Tables~\ref{tab:TAB_adult} - \ref{tab:TAB_spambase}).
As we can see in \tablename~\ref{tab:TABELLONE_RISULTATI}, NAIM achieves the best performance, highlighted in bold, in most scenarios (23 out of 36).
To be more specific, we conducted the statistical analysis described in section~\ref{sec:metrics_stats} and reported in~\ref{app:stats}, to compare the predictions made by NAIM in comparison with those of its competitors.
This analysis shows that on average the proposed model achieves better performance in $58.7\%$ of the cases, while it only loses in $1.6\%$ of the cases.

\begin{figure*}[!ht]
    \centering
    \resizebox{\textwidth}{!}{
    \begin{tikzpicture}
    \node[label={[xshift=.5cm]left:\scriptsize{(A)}}] (ML) at (0,0) {\includegraphics[width=\textwidth]{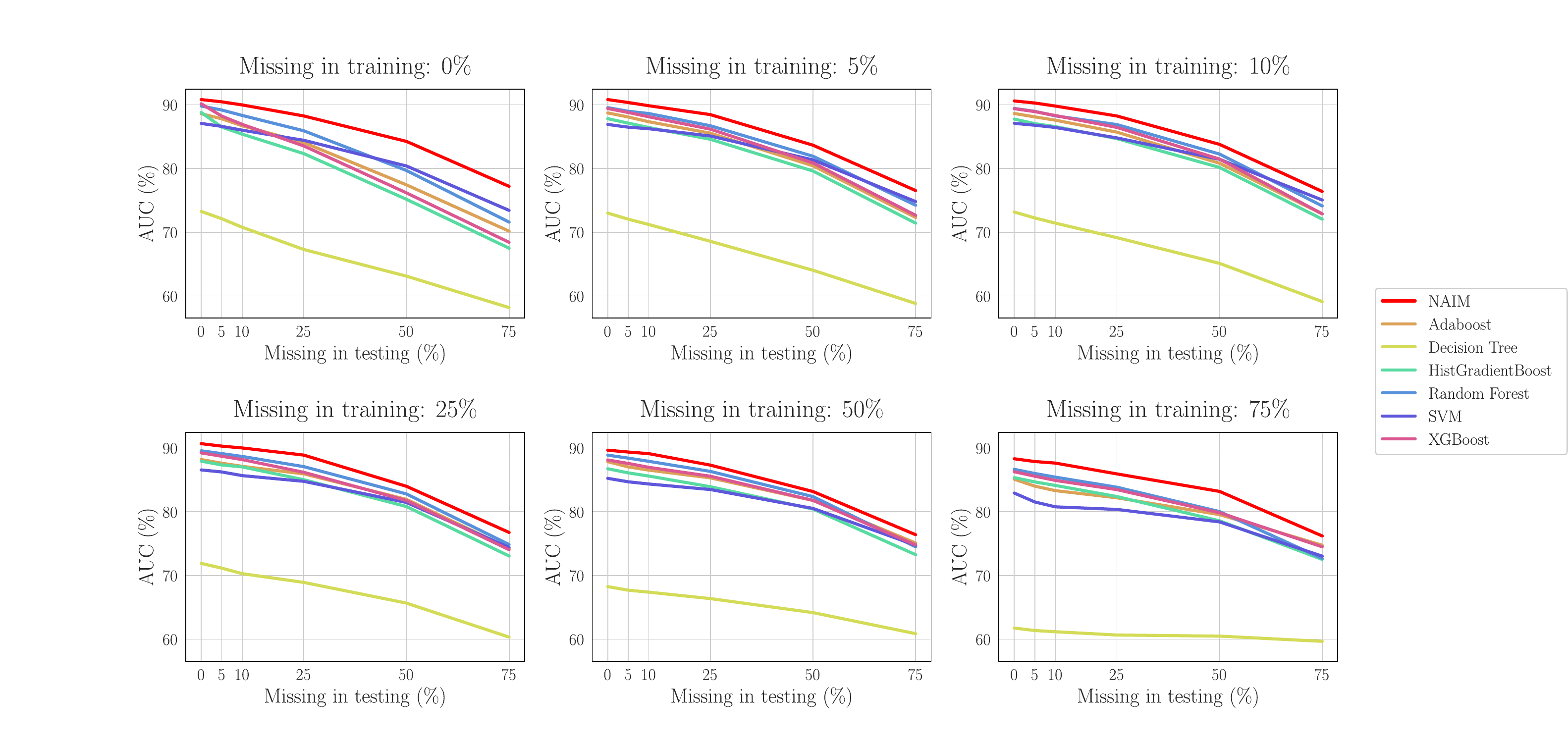}};
    \node[label={[xshift=.5cm]left:\scriptsize{(B)}}] (DL) [below left=-.2cm and -9cm of ML.south] {\includegraphics[width=.99\textwidth]{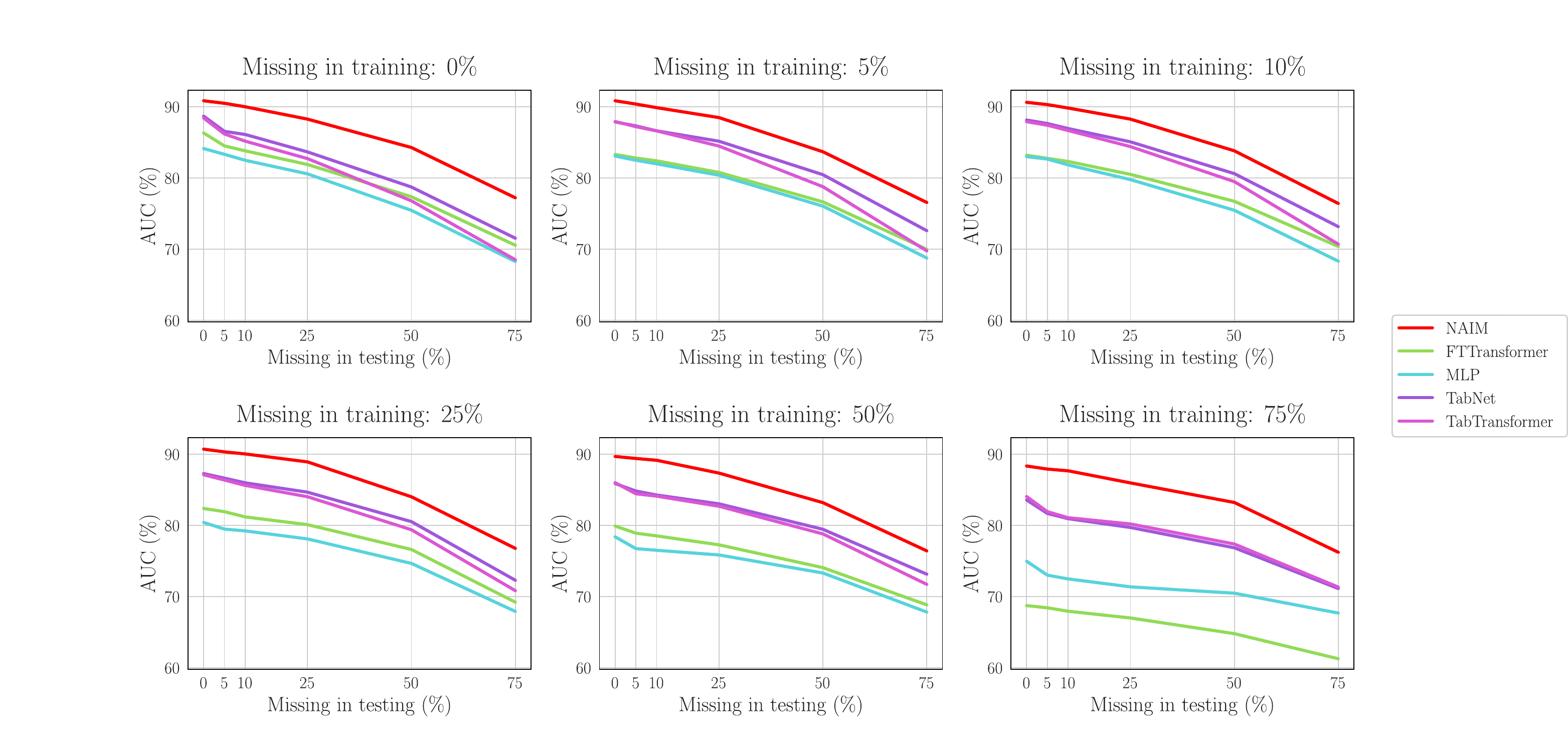}};
    \draw[-, thick] ([xshift=-.3cm]ML.south west) -- (ML.south east);
    \end{tikzpicture}}
    \caption{Comparison between the test performance of NAIM and the competitor models averaged over the $5$ dataset and the $3$ imputation strategies: panel A reports the comparison between NAIM and the ML models, whereas panel B compares NAIM with the DL models.}
    \label{fig:naim_vs_ML_DL}
\end{figure*}

\begin{figure*}[!ht]
    \centering
    \resizebox{\textwidth}{!}{
    \begin{tikzpicture}
        \node (fig) {\includegraphics[width=\textwidth]{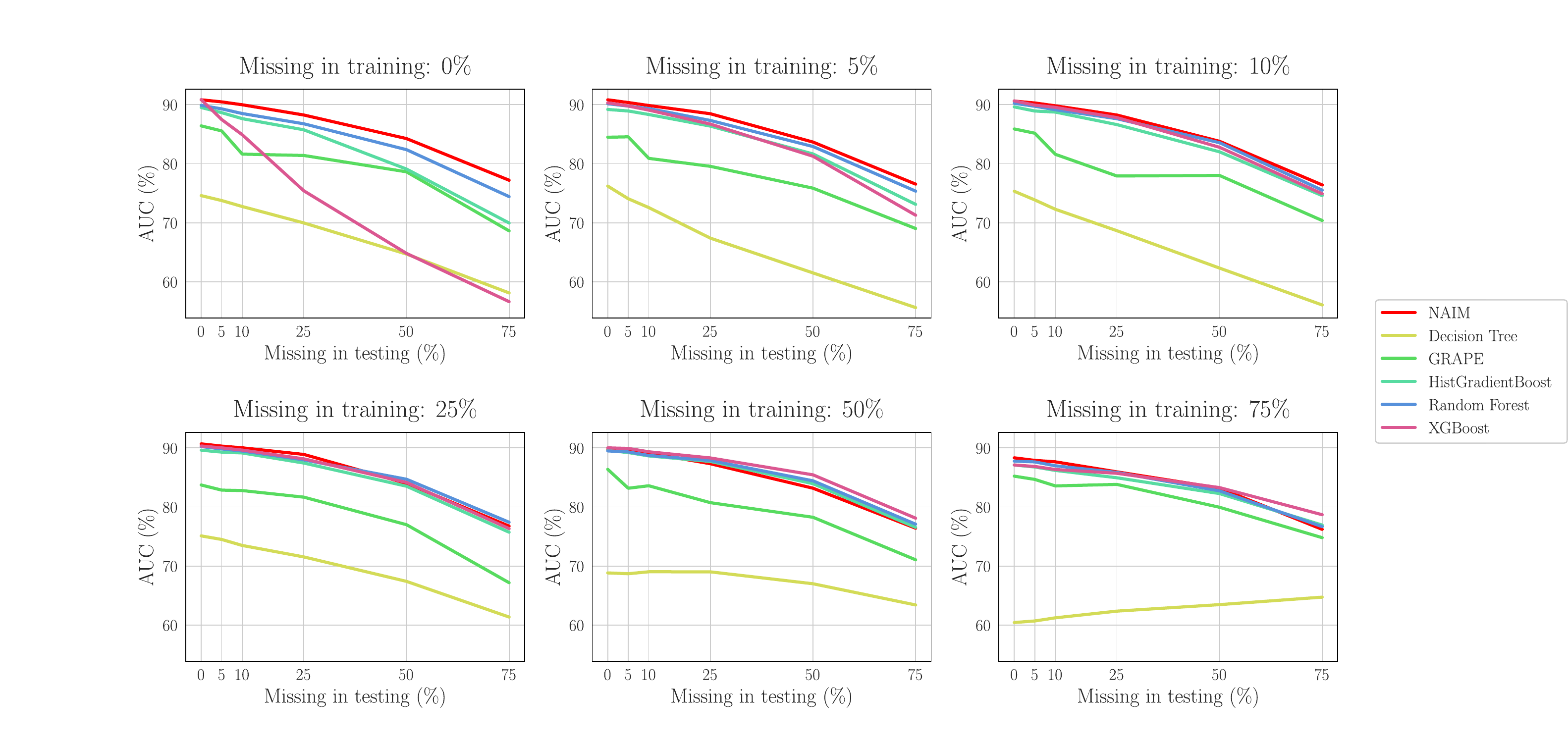}};
        \draw[-, thick, draw=white] ([xshift=-.3cm]fig.south west) -- (fig.south east);
    \end{tikzpicture}}
    \caption{Comparison between the test performance of NAIM and the competitors implementing an intrinsic strategy averaged over the $5$ dataset.}
    \label{fig:naim_vs_MIA}
\end{figure*}

\begin{figure*}[!ht]
    \centering
    \resizebox{\textwidth}{!}{
    \begin{tikzpicture}
        \node (fig) {\includegraphics[width=\textwidth]{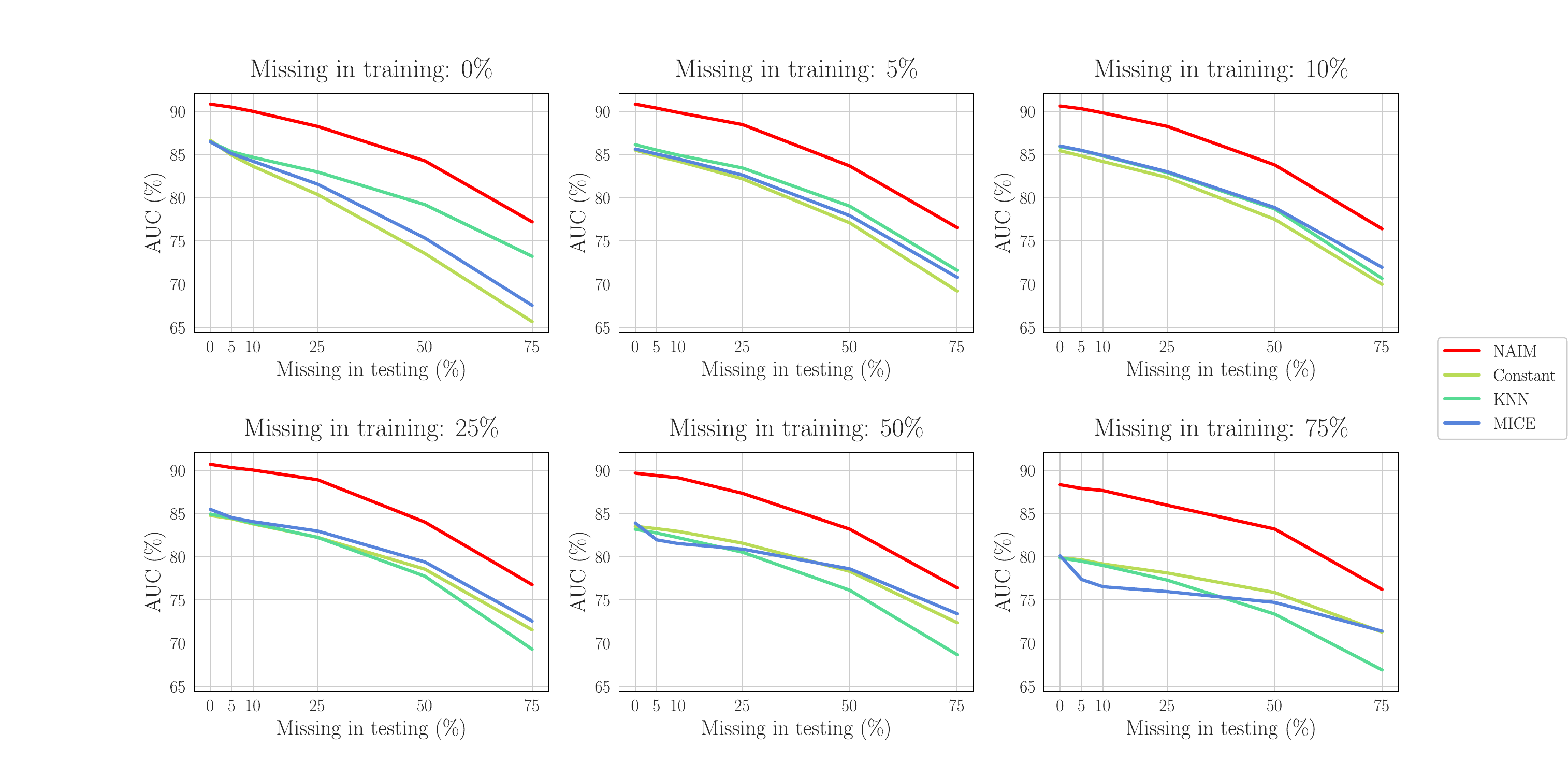}}; 
        \draw[-, thick, draw=white] ([xshift=-.3cm]fig.south west) -- ([xshift=.8cm]fig.south east);
    \end{tikzpicture}}
    \caption{Comparison between the test performance of NAIM and the competitor imputers averaged over the $5$ dataset and the $10$ models available.}
    \label{fig:naim_vs_imputation}
\end{figure*}

To further analyze the results we divide the competitors into groups and plot the performance constructing $6$ separate charts, one for each level of missing data within the training set, which report the performance metrics as the percentage of missing data in the test set increases.
We use this representation to compare NAIM with different groups of competitors, namely ML and DL models paired with the imputation techniques (\figurename~\ref{fig:naim_vs_ML_DL}), model implementing some intrinsic strategy (\figurename~\ref{fig:naim_vs_MIA}) and the various imputation techniques regardless of the model used (\figurename~\ref{fig:naim_vs_imputation}).
To present our findings, these charts report the average performance values. 

As a first analysis, we compare NAIM with ML and DL models coupled with the 3 different imputers under consideration. 
In \figurename~\ref{fig:naim_vs_ML_DL}, in panels A and B for the ML and DL models respectively, we reported the average performance across all $5$ datasets and the $3$ imputation techniques under consideration.
These charts show the decreasing trend in model's performance as the percentage of missing values in the testing set increases, a finding that aligns with our expectations.
Interestingly, NAIM stands out for its consistently superior average performance across all levels of missing data, maintaining its high ranking even in the optimal scenario where no data is missing. 
This observation not only underscores the distinct advantage of NAIM over ML and DL models but also highlights the unexplored potential of DL methods in improving both the model's performance when handling missing data.
Furthermore, the performance difference is particularly evident when the training set has $0\%$ missing data (first chart on the left in the first row for each panel A and B). 
This highlights a significant challenge that current methodologies face: the need to learn how to handle missing data during the training phase, to then correctly infer on testing data with missing values. 
This aspect emphasizes a critical limitation of conventional approaches and the need for innovative strategies that can effectively address data incompleteness.
In this respect, we deem that the better performance of NAIM is due to our regularization technique that enables it to learn how to handle missing values even if all the training data are present.

Then, as a second analysis, we compare NAIM against the models capable of intrinsically handling missing values.
In \figurename~\ref{fig:naim_vs_MIA}, which reports the performance of these approaches, the average performances are computed across the $5$ datasets.
Turning our focus on the top left chart, which plots the performance in the $0\%$ missing values in training scenario, we notice that the difference in performance as the percentage of missing values in the test set increases is higher compared to the other $5$ charts where the training is performed on data containing missing values. 
This performance drop confirms that neither the imputation techniques nor the intrinsic strategies are effective if no missing values have been seen during training, whereas the stable performances of NAIM validate our novel regularization technique, which allows it to adapt to the presence of missing data in all training scenarios.
These charts also show the competitive performance obtained by XGBoost and Random Forest, each delivering comparable performance across a spectrum of scenarios characterized by missing data in the training set, including the optimal scenario of $0\%$ missing in both sets. 
It is worth noting that in the scenario with $50\%$ missing data in training (second row, second chart from the left), they are even able to outperform NAIM regardless of the percentage of missing values in the test set. 
Moreover, it is noteworthy that HistGradientBoost achieves performance similar to those of XGBoost and Random Forest, although slightly inferior. 
\rr On the opposite, Decision Tree and GRAPE, despite being theoretically capable of handling missing data, consistently fail to achieve comparable performance to other models in both missing data scenarios during training and testing. \bb

Subsequently, as a third analysis, to further assess NAIM's ability to handle missing data, we compare it against imputation strategies.
In contrast to the previous analyses, here the average performance for each of the imputation methods, reported in \figurename~\ref{fig:naim_vs_imputation}, are computed by averaging the results across the $5$ datasets and the $10$ models available.
The figure illustrates how, in scenarios involving complete training data (first row, first chart from the left), models trained with data imputed using the KNN imputer exhibit slightly superior performance compared to other imputation methods. 
However, it is important to note that, overall, there are no significant and clear differences among the various imputation techniques, thus requiring extensive experimentation to find the optimal configuration for the data and task at hand.

\begin{figure}[!ht]
    \centering
    \includegraphics[width=\columnwidth]{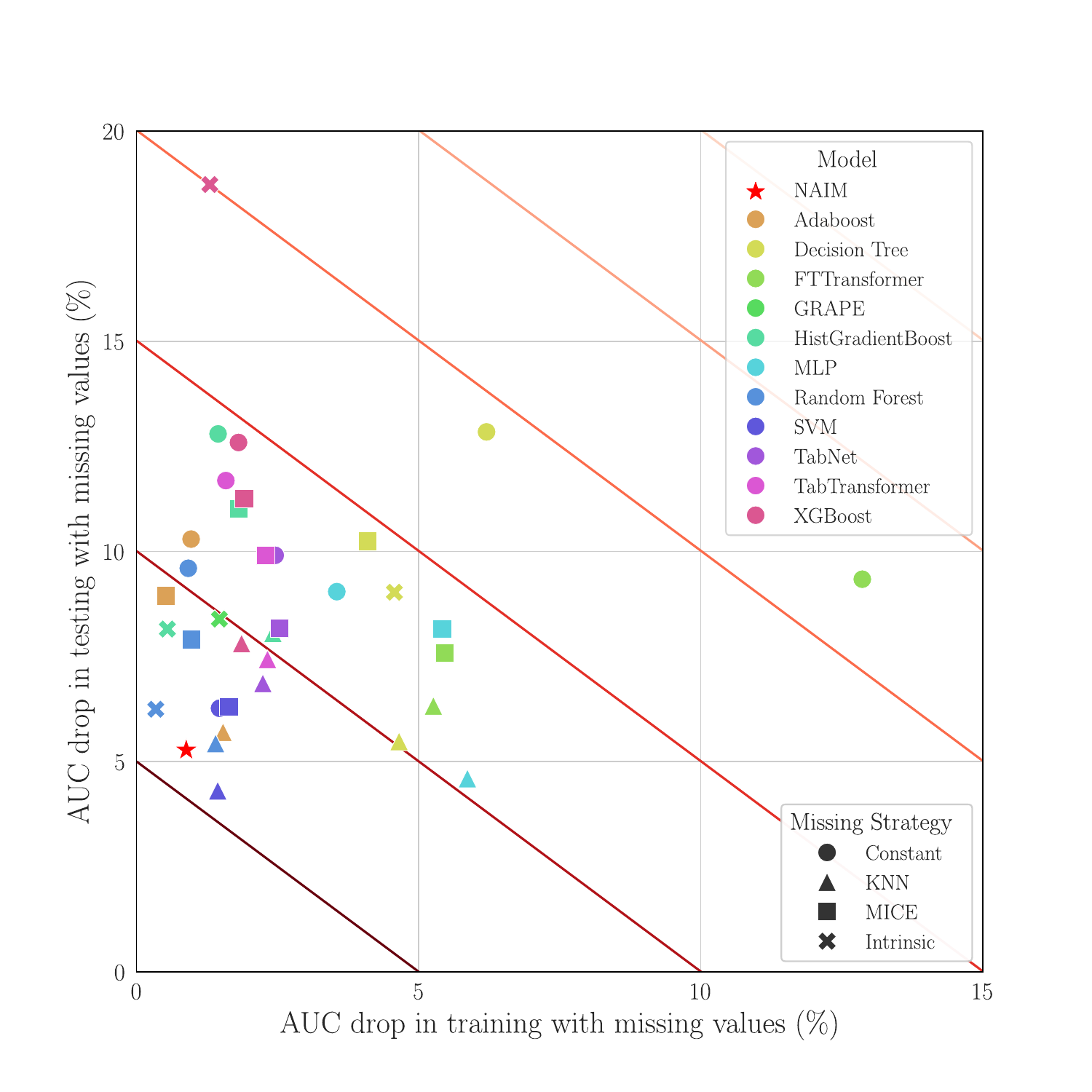}
    \caption{Analysis of the robustness of the models to increasing levels of missing data. Each experiment (combination of model and missing strategy) is positioned based on the average drop in performance obtained in scenarios with increasing percentages of missing values in the training and testing sets, separately.}
    \label{fig:missing_robustness}
\end{figure}

Upon observing the superior performance of NAIM across all missing data scenarios, we examine its performance robustness to varying levels of missing data (\figurename~\ref{fig:missing_robustness}). 
To this end, we measure the performance drop presented in experiments with missing data either in the training or in the test set compared to the optimal scenario of $0\%$ of missing data in training and test sets.
Subsequently, we average these values to present them in a single plot.
To quantify the robustness of the models when dealing with increasing missing values in testing, we evaluate the performance differences across scenarios with complete training data and various missing data percentages in the test sets.
In other words, to compute the $y$ coordinates of \figurename~\ref{fig:missing_robustness}, we measure the average performance drop in the experiments reported in the first chart of the first row of panels A and B of \figurename~\ref{fig:naim_vs_ML_DL} and in \figurename~\ref{fig:naim_vs_MIA}.
Conversely, to assess the robustness of the models when dealing with increasing missing values in training, we examine the performance differences in scenarios with various missing data percentages in training and complete test data. 
Therefore, we measure the differences among the first performance reported in each of the $6$ charts shown in panels A and B of \figurename~\ref{fig:naim_vs_ML_DL} and in \figurename~\ref{fig:naim_vs_MIA} and report their average as $x$ coordinates in \figurename~\ref{fig:missing_robustness}.
To facilitate interpretation and effectively rank models based on their robustness, we also plot equi-drop lines connecting data points that exhibit equivalent overall performance reductions from the no-missing data state ($0\%$ of missing values in both sets).
For instance, an experiment exhibiting an average $10\%$ drop in the presence of missing data in the test set and $0\%$ drop in the training set, would be ranked equivalently to a model with the opposite behavior of $0\%$ drop in the presence of missing data in the training set and $10\%$ drop in the set set.
Straightforwardly, a model completely robust to missing data in both training and testing sets would be placed on the point $(0,0)$.
We note that NAIM's robustness to missing data is indicative of minimal performance degradation with an increasing number of missing values, since it presents a $0.88\%$ drop when the rate of missing values increases in the training set and a $5.27\%$ drop when the rate of missing values increases in the test set. 
Notably, despite this analysis positions NAIM just behind the SVM model paired with the KNN imputer, when considering the comprehensive performance evaluations presented in panel A of \figurename~\ref{fig:naim_vs_ML_DL} and in \tablename~\ref{tab:TABELLONE_RISULTATI}, NAIM significantly surpasses the competitor.
Furthermore, looking at \tablename~\ref{tab:TABELLONE}, we can see that NAIM performance is statistically larger in $67.8\%$ of the cases, whereas it is never inferior.
This means that SVM is a more robust model but with lower performance compared to NAIM, confirming that our approach presents a superior capacity to maintain high-performance levels despite the presence of missing data. 
Moreover, from \figurename~\ref{fig:missing_robustness} we can see that the third model in terms of robustness in the presence of missing values is the Random Forest that exploits the MIA strategy, which also performs statistically worse than NAIM as depicted in \tablename~\ref{tab:TABELLONE}. 
These observations confirm the robustness of the proposed model and regularization technique in handling missing values.

\begin{figure*}[!ht]
    \centering
    \resizebox{\textwidth}{!}{
    \begin{tikzpicture}
        \node (fig) {\includegraphics[width=3.\textwidth]{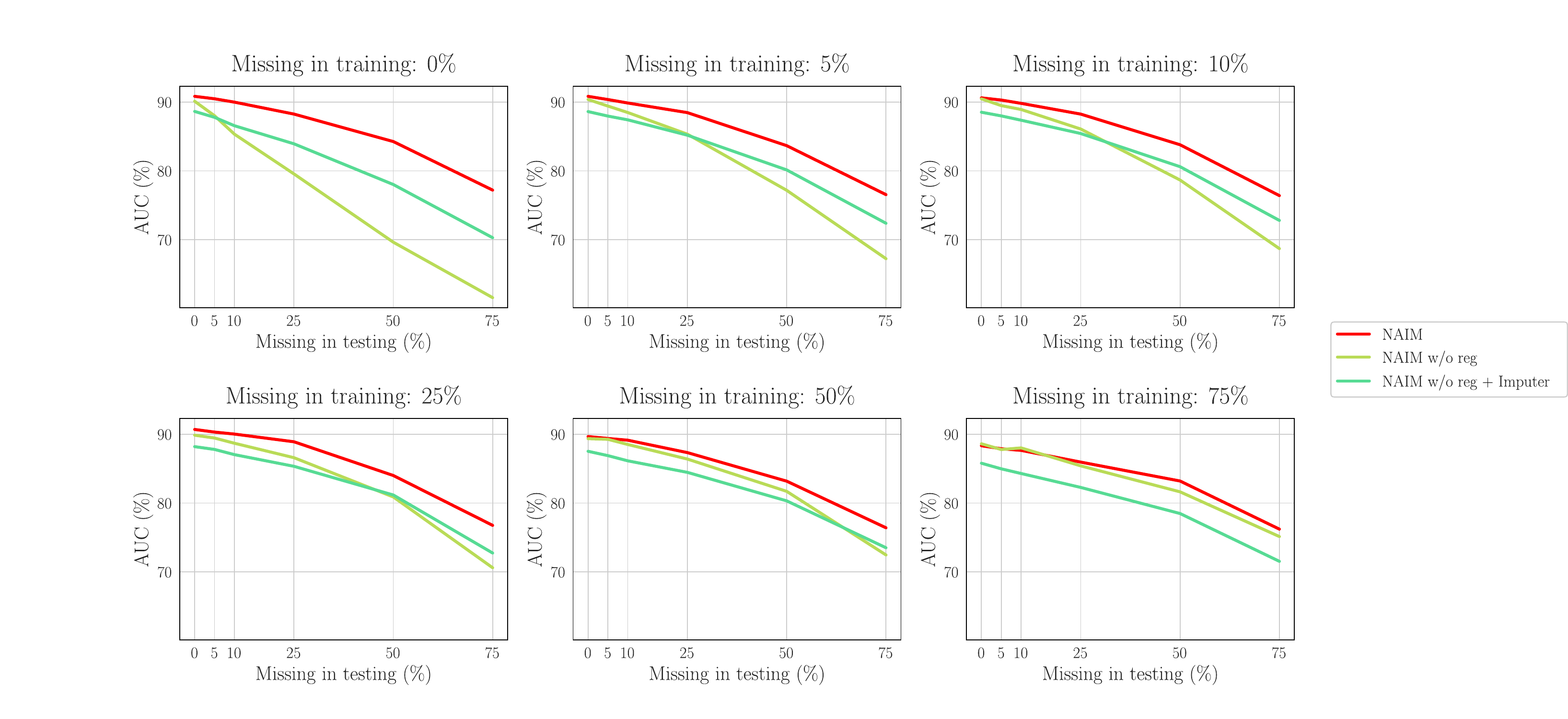}}; 
        \draw[-, thick, draw=white] ([xshift=-1.5cm]fig.south west) -- ([xshift=-2cm]fig.south east);
    \end{tikzpicture}}
    \caption{\rr Ablation analysis evaluating the impact of the regularization technique (NAIM w/o reg) and the masked self-attention mechanism (NAIM w/o reg + Imputer). The plots report test performance averaged over the $5$ datasets. For NAIM w/o reg + Imputer, results are additionally averaged across the $3$ different imputers.\bb}
    \label{fig:ablation}
\end{figure*}

\rr 
As a further analysis to evaluate the contributions of the regularization technique and the masking mechanism, we conducted ablation studies in which the model was trained without regularization, both in scenarios with missing values (NAIM w/o reg) and when combined with imputation techniques (NAIM w/o reg + Imputer).
\figurename~\ref{fig:ablation} shows the average performance of these experiments conducted across the five datasets, while a comprehensive presentation of the results is provided in Tables~\ref{tab:TABELLONE_RISULTATI}–\ref{tab:TAB_spambase} in~\ref{app:results}.
Notably, the contribution of the regularization technique is evident throughout the $6$ plots. 
Indeed, in the first plot (top left), NAIM w/o reg exhibits a performance trend comparable to that of XGBoost in \figurename~\ref{fig:naim_vs_MIA}, suggesting that exposure to missing data during training is essential for the model to learn effectively how to handle them.
Conversely, in the final plot (bottom right), where the missing rate is highest, the benefit of regularization becomes less pronounced.
When evaluating NAIM combined with the three imputation strategies, it is evident that the masking self-attention mechanism consistently outperforms the imputation-based baselines across all epochs.
These findings indicate that each component of our approach, particularly the missing data regularization, contributes significantly to the overall performance of NAIM.
\bb

\section{Conclusion}\label{sec:conclusion}

In this work, we introduced NAIM, an innovative transformer-based architecture for modeling tabular data in supervised learning environments, specifically designed to handle missing values. 
This architecture simplifies the analysis process, essentially avoiding the need for traditional imputation strategies. 
Our empirical analysis demonstrates NAIM's superiority over existing state-of-the-art solutions in handling missing data within tabular datasets. 
Our experiments spanned across NAIM and a combination of $11$ ML and DL models, each integrated with $3$ distinct imputation methods, and also against those $5$ models that intrinsically handle missing values.
We performed the analyses on $5$ publicly available datasets, which differ in the number of samples and features.

NAIM leverages specialized embedding mechanisms for both categorical and numerical features, coupled with an innovative self-attention mechanism, to maximize the information of the available data at hand.
Additionally, we introduced a novel regularization strategy designed to address a significant challenge identified in our research: the need for state-of-the-art approaches to learn how to handle missing values in the training process. 
This strategy, which involves random masking of each sample in every epoch, equips NAIM with the robust capability to handle missing data under all circumstances, including scenarios where there are no missing values in the training set.

\rr 
While we have attained promising results, this work has some limitations that are now discussed introducing the related future directions:

\begin{itemize}
\item Multimodal learning with missing modalities: We plan to extend NAIM beyond the tabular domain by developing a multimodal variant capable of handling heterogeneous data types and the complete absence of entire modalities (e.g., missing imaging or tabular information). 
This extension would enable a unified framework that leverages available data across different domains while  degrading as little as possible in performance when some sources are absent.
\item Efficient attention mechanisms: Although transformers are powerful, they are known for their high computational cost. 
Future investigations should explore alternative attention formulations that can reduce memory and computational requirements. 
This would make NAIM more scalable, particularly for large datasets with high-dimensional features.
\item Model interpretability and transparency: To foster trust and adoption in critical domains like healthcare and finance, future studies should deepen the interpretability of NAIM. 
This includes leveraging attention weights to understand feature importance, analyzing learned embeddings to uncover latent data structure~\cite{bib:embed_space}, and developing attribution methods tailored to transformers with missing inputs. 
These efforts would help demystify the internal reasoning process of the model, especially in the presence of incomplete data.
\item Self-supervised and semi-supervised learning: Since labeled tabular data is often scarce, particularly in specialized domains, future works should investigate self-supervised pretraining strategies for NAIM. 
Techniques such as contrastive learning, masked feature modeling, or denoising objectives could help the model learn robust representations from raw data, improving downstream performance when only limited supervision is available.
\item Temporal modeling for incomplete time-series: Another natural extension of our framework involves adapting NAIM to handle longitudinal or sequential data.
Time-series datasets, especially in domains like sensor monitoring or electronic health records, frequently exhibit irregular sampling and missingness. 
Future work should aim to develop a time-aware version of NAIM that integrates temporal encodings.
\item Domain-specific evaluation and adaptation: While our evaluation included diverse public datasets, future work should assess NAIM in domain-specific real-world applications. 
Particular attention should be given to healthcare datasets, where missingness is often informative and non-random. 
This would not only validate the model’s utility in complex scenarios but also drive the development of domain-adaptive mechanisms that incorporate domain knowledge into the learning process.
\end{itemize}
\bb

These potential directions not only underscore the versatility and expansiveness of our approach but also highlight the fertile ground for future advancements in the domain of tabular data analysis.
In conclusion, when faced with tabular datasets exhibiting missing data either in the training or testing phase, the simplest and most reliable choice is NAIM, as it enables the extraction of useful information from the available data without any need for imputation.

\section*{Acknowledgment}
Camillo Maria Caruso is a Ph.D. student enrolled in the National Ph.D. in Artificial Intelligence, XXXVII cycle, course on Health and life sciences, organized by Università Campus Bio-Medico di Roma.

This work was partially founded by: 
i) Università Campus Bio-Medico di Roma under the program ``University Strategic Projects'' within the project ``AI-powered Digital Twin for next-generation lung cancEr cAre (IDEA)''; 
ii) from PRIN 2022 MUR 20228MZFAA-AIDA (CUP C53D23003620008); 
iii) from PRIN PNRR 2022 MUR P2022P3CXJ-PICTURE (CUP C53D23009280001);
iv) from PNRR MUR project PE0000013-FAIR.

Resources are provided by the National Academic Infrastructure for Supercomputing in Sweden (NAISS) and the Swedish National Infrastructure for Computing (SNIC) at Alvis @ C3SE, partially funded by the Swedish Research Council through grant agreements no. 2022-06725 and no. 2018-05973.

\section*{Author Contributions}
\textbf{Camillo Maria Caruso:} Conceptualization, Data curation, Formal analysis, Investigation, Methodology, Software, Validation, Visualization, Writing – original draft, Writing – review \& editing;
\textbf{Paolo Soda:} Conceptualization, Funding acquisition, Investigation, Methodology, Project administration, Resources, Supervision, Writing – review \& editing;
\textbf{Valerio Guarrasi:} Conceptualization, Formal analysis, Funding acquisition, Investigation, Methodology, Project administration, Resources, Supervision, Validation, Visualization, Writing – original draft, Writing – review \& editing.



\bibliographystyle{IEEEtran} 
\bibliography{bibliography.bib}

\vspace{11pt}

\appendix
\clearpage
\onecolumn

\section{Masking example}\label{app:mask}

In this section, we aim to illustrate in more depth how the proposed masked self-attention mechanism is able to mask and ignore the contribution of missing values via an example.

Let the feature vector $x$ be composed of 4 different features, out of which the third feature is missing:
\begin{equation}
\scalebox{.75}{$
x = \renewcommand\arraystretch{1.5}\begin{bmatrix}[*4{C{6.5mm}}]
    $x_1$ \\
    $x_2$ \\
    $x_3$ \\
    $x_4$ \\
    \end{bmatrix}$}
\end{equation}

Let us now compute the embedding matrix $e$ of the features, using equations~\ref{eq:cat_emb} and \ref{eq:num_emb} for the categorical and numerical features respectively, using a dimension of the embeddings $d_e = 4$: 
\begin{equation}
\scalebox{.75}{$
e = \renewcommand\arraystretch{1.5}\begin{bmatrix}[C{6.5mm}]
    $a$ \\
    $b$ \\
    $c$ \\
    $d$ \\
    \end{bmatrix} = 
    \renewcommand\arraystretch{1.5}\begin{bmatrix}[*4{C{6.5mm}}]
    $a_1$ &          $a_2$ &        $a_3$ &        $a_4$ \\
    $b_1$ &          $b_2$ &        $b_3$ &        $b_4$ \\
    $c_1$ &          $c_2$ &        $c_3$ &        $c_4$ \\
    $d_1$ &          $d_2$ &        $d_3$ &        $d_4$ \\
    \end{bmatrix}$}
\end{equation}

As $x_3$ was missing, we can remark that embedding $c$ denotes the missing feature and the mask $M$ can be defined:
\begin{equation}
\scalebox{.75}{$
M = \renewcommand\arraystretch{1.5}\begin{bmatrix}[*4{C{6.7mm}}]
                0 &             0 &           -$\infty$  & 0\\
                0 &             0 &           -$\infty$  & 0 \\
                0 &             0 &           -$\infty$ & 0 \\
                0 &             0 &           -$\infty$ & 0 \\
                \end{bmatrix}
$}
\end{equation}

Then, following the steps in the transformer pipeline, we compute $Q$, $K$ and $V$ using Eq.~\ref{eq:embeddings} and a number of heads $h = 2$:
\begin{equation}
\scalebox{.75}{$
\begin{aligned}
Q = \renewcommand\arraystretch{1.5}\begin{bmatrix}[C{6.5mm}]
    $a^Q$ \\
    $b^Q$ \\
    $c^Q$ \\
    $d^Q$ \\
    \end{bmatrix} = 
    \renewcommand\arraystretch{1.5}\begin{bmatrix}[*2{C{6.5mm}}]
    $a_1^Q$ &          $a_2^Q$ \\
    $b_1^Q$ &          $b_2^Q$ \\
    $c_1^Q$ &          $c_2^Q$ \\
    $d_1^Q$ &          $d_2^Q$ \\
    \end{bmatrix}, &&
K = \renewcommand\arraystretch{1.5}\begin{bmatrix}[C{6.5mm}]
    $a^K$ \\
    $b^K$ \\
    $c^K$ \\
    $d^K$ \\
    \end{bmatrix} = 
    \renewcommand\arraystretch{1.5}\begin{bmatrix}[*2{C{6.5mm}}]
    $a_1^K$ &          $a_2^K$ \\
    $b_1^K$ &          $b_2^K$ \\
    $c_1^K$ &          $c_2^K$ \\
    $d_1^K$ &          $d_2^K$ \\
    \end{bmatrix}, &&
V = \renewcommand\arraystretch{1.5}\begin{bmatrix}[C{6.5mm}]
    $a^V$ \\
    $b^V$ \\
    $c^V$ \\
    $d^V$ \\
    \end{bmatrix} = 
    \renewcommand\arraystretch{1.5}\begin{bmatrix}[*2{C{6.5mm}}]
    $a_1^V$ &          $a_2^V$ \\
    $b_1^V$ &          $b_2^V$ \\
    $c_1^V$ &          $c_2^V$ \\
    $d_1^V$ &          $d_2^V$ \\
    \end{bmatrix}
\end{aligned}
$}
\end{equation}

We now compute the $QK^T$ product to better understand how the contributions from various features ($a$, $b$, $c$ and $d$) are distributed across the attention matrix.
\begin{equation}
\scalebox{.75}{$
QK^T = \renewcommand\arraystretch{1.5}\begin{bmatrix}[*4{C{13mm}}]
    $a^Q a^K$ &          $a^Q b^K$      &       $a^Q c^K$ &          $a^Q d^K$ \\
    $b^Q a^K$ &          $b^Q b^K$      &       $b^Q c^K$ &          $b^Q d^K$ \\
    $c^Q a^K$ &          $c^Q b^K$      &       $c^Q c^K$ &          $c^Q d^K$ \\
    $d^Q a^K$ &          $d^Q b^K$      &       $d^Q c^K$ &          $d^Q d^K$ \\
    \end{bmatrix}
$}
\end{equation}

Now, using the $QK^T$ product, the vector $V$ and the mask $M$, we can calculate the output of an attention head for the classical masked self-attention mechanism applied to tabular data:
\begin{equation}
\resizebox{.88\textwidth}{!}{$
\begin{aligned}
&\mathit{softmax}(QK^T + M)V =  \\ 
& = \mathit{softmax}\left(\renewcommand\arraystretch{1.5}\begin{bmatrix}[*4{C{13mm}}]
    $a^Q a^K$ &          $a^Q b^K$      &       $a^Q c^K$ &          $a^Q d^K$ \\
    $b^Q a^K$ &          $b^Q b^K$      &       $b^Q c^K$ &          $b^Q d^K$ \\
    $c^Q a^K$ &          $c^Q b^K$      &       $c^Q c^K$ &          $c^Q d^K$ \\
    $d^Q a^K$ &          $d^Q b^K$      &       $d^Q c^K$ &          $d^Q d^K$ \\
    \end{bmatrix} + 
    \renewcommand\arraystretch{1.5}\begin{bmatrix}[*4{C{6mm}}]
                0 &             0 &           -$\infty$  & 0\\
                0 &             0 &           -$\infty$  & 0 \\
                0 &             0 &           -$\infty$ & 0 \\
                0 &             0 &           -$\infty$ & 0 \\
                \end{bmatrix} \right)
    \renewcommand\arraystretch{1.5}\begin{bmatrix}[*2{C{6mm}}]
    $a_1^V$ &          $a_2^V$ \\
    $b_1^V$ &          $b_2^V$ \\
    $c_1^V$ &          $c_2^V$ \\
    $d_1^V$ &          $d_2^V$ \\
    \end{bmatrix} = \\
& = \renewcommand\arraystretch{1.5}\begin{bmatrix}[C{26mm}*3{C{17mm}}] 
    $\mathit{softmax}(a^Q a^K$ &          $a^Q b^K$      &       -$\infty$ &          $a^Q d^K)$ \\
    $\mathit{softmax}(b^Q a^K$ &          $b^Q b^K$      &       -$\infty$ &          $b^Q d^K)$ \\
    $\mathit{softmax}(c^Q a^K$ &          $c^Q b^K$      &       -$\infty$ &          $c^Q d^K)$ \\
    $\mathit{softmax}(d^Q a^K$ &          $d^Q b^K$      &       -$\infty$ &          $d^Q d^K)$ \\
    \end{bmatrix}
    \renewcommand\arraystretch{1.5}\begin{bmatrix}[*2{C{6mm}}]
    $a_1^V$ &          $a_2^V$ \\
    $b_1^V$ &          $b_2^V$ \\
    $c_1^V$ &          $c_2^V$ \\
    $d_1^V$ &          $d_2^V$ \\
    \end{bmatrix} = \label{eq:classical_msa} \\
    & = \renewcommand\arraystretch{1.5}\begin{bmatrix}[C{26mm}*3{C{17mm}}] 
    $(a^Q a^K)_S$ &          $(a^Q b^K)_S$      &       0 &          $(a^Q d^K)_S$ \\
    $(b^Q a^K)_S$ &          $(b^Q b^K)_S$      &       0 &          $(b^Q d^K)_S$ \\
    $(c^Q a^K)_S$ &          $(c^Q b^K)_S$      &       0 &          $(c^Q d^K)_S$ \\
    $(d^Q a^K)_S$ &          $(d^Q b^K)_S$      &       0 &          $(d^Q d^K)_S$ \\
    \end{bmatrix} 
    \renewcommand\arraystretch{1.5}\begin{bmatrix}[*2{C{6mm}}]
    $a_1^V$ &          $a_2^V$ \\
    $b_1^V$ &          $b_2^V$ \\
    $c_1^V$ &          $c_2^V$ \\
    $d_1^V$ &          $d_2^V$ \\
    \end{bmatrix} =  \\ 
    &= \renewcommand\arraystretch{1.5}\begin{bmatrix}[*2{C{80mm}}]
    $(a^Q a^K)_S \cdot a_1^V + (a^Q b^K)_S \cdot b_1^V + (a^Q d^K)_S \cdot d_1^V + 0$ & $(a^Q a^K)_S \cdot a_2^V + (a^Q b^K)_S \cdot b_2^V + (a^Q d^K)_S \cdot d_2^V + 0$ \\
    $(b^Q a^K)_S \cdot a_1^V + (b^Q b^K)_S \cdot b_1^V + (b^Q d^K)_S \cdot d_1^V + 0$ & $(b^Q a^K)_S \cdot a_2^V + (b^Q b^K)_S \cdot b_2^V + (b^Q d^K)_S \cdot d_2^V + 0$ \\
    $(c^Q a^K)_S \cdot a_1^V + (c^Q b^K)_S \cdot b_1^V + (c^Q d^K)_S \cdot d_1^V + 0$ & $(c^Q a^K)_S \cdot a_2^V + (c^Q b^K)_S \cdot b_2^V + (c^Q d^K)_S \cdot d_2^V + 0$ \\
    $(d^Q a^K)_S \cdot a_1^V + (d^Q b^K)_S \cdot b_1^V + (d^Q d^K)_S \cdot d_1^V + 0$ & $(d^Q a^K)_S \cdot a_2^V + (d^Q b^K)_S \cdot b_2^V + (d^Q d^K)_S \cdot d_2^V + 0$ \\
    \end{bmatrix} 
\end{aligned}$}
\end{equation}

As can be seen, some contributions of the missing feature are still present on the third line of the output matrix.
Using as a starting point, for the sake of simplicity, the attention matrix computed on the third step of the Eq.~\ref{eq:classical_msa}, let us compute the output matrix obtained using the presented masked self-attention mechanism:
\begin{equation}
\resizebox{.88\textwidth}{!}{$
\begin{aligned}
&\mathit{ReLU}(\mathit{softmax}(QK^T + M) + M^T)V = \\
&= \mathit{ReLU}\left(\renewcommand\arraystretch{1.5}\begin{bmatrix}[*4{C{18.3mm}}] 
    $(a^Q a^K)_S$ &          $(a^Q b^K)_S$      &       0 &          $(a^Q d^K)_S$ \\
    $(b^Q a^K)_S$ &          $(b^Q b^K)_S$      &       0 &          $(b^Q d^K)_S$ \\
    $(c^Q a^K)_S$ &          $(c^Q b^K)_S$      &       0 &          $(c^Q d^K)_S$ \\
    $(d^Q a^K)_S$ &          $(d^Q b^K)_S$      &       0 &          $(d^Q d^K)_S$ \\
    \end{bmatrix} + 
    \renewcommand\arraystretch{1.5}\begin{bmatrix}[*4{C{6mm}}]
                0 &             0 &           0  & 0\\
                0 &             0 &           0  & 0 \\
        -$\infty$ &     -$\infty$ &    -$\infty$ & -$\infty$ \\
                0 &             0 &           0 & 0 \\
                \end{bmatrix} \right)
    \renewcommand\arraystretch{1.5}\begin{bmatrix}[*2{C{6mm}}]
    $a_1^V$ &          $a_2^V$ \\
    $b_1^V$ &          $b_2^V$ \\
    $c_1^V$ &          $c_2^V$ \\
    $d_1^V$ &          $d_2^V$ \\
    \end{bmatrix} = \\
    &= \mathit{ReLU}\left(\renewcommand\arraystretch{1.5}\begin{bmatrix}[*4{C{18.3mm}}] 
    $(a^Q a^K)_S$ &          $(a^Q b^K)_S$      &       0 &          $(a^Q d^K)_S$ \\
    $(b^Q a^K)_S$ &          $(b^Q b^K)_S$      &       0 &          $(b^Q d^K)_S$ \\
                 -$\infty$ &     -$\infty$ &    -$\infty$ & -$\infty$ \\
    $(d^Q a^K)_S$ &          $(d^Q b^K)_S$      &       0 &          $(d^Q d^K)_S$ \\
    \end{bmatrix}\right)
    \renewcommand\arraystretch{1.5}\begin{bmatrix}[*2{C{6mm}}]
    $a_1^V$ &          $a_2^V$ \\
    $b_1^V$ &          $b_2^V$ \\
    $c_1^V$ &          $c_2^V$ \\
    $d_1^V$ &          $d_2^V$ \\
    \end{bmatrix} = \\
    &= \renewcommand\arraystretch{1.5}\begin{bmatrix}[*4{C{18.3mm}}] 
    $(a^Q a^K)_S$ &          $(a^Q b^K)_S$      &       0 &          $(a^Q d^K)_S$ \\
    $(b^Q a^K)_S$ &          $(b^Q b^K)_S$      &       0 &          $(b^Q d^K)_S$ \\
                0 &                      0      &       0 &                      0 \\
    $(d^Q a^K)_S$ &          $(d^Q b^K)_S$      &       0 &          $(d^Q d^K)_S$ \\
    \end{bmatrix}
    \renewcommand\arraystretch{1.5}\begin{bmatrix}[*2{C{6mm}}]
    $a_1^V$ &          $a_2^V$ \\
    $b_1^V$ &          $b_2^V$ \\
    $c_1^V$ &          $c_2^V$ \\
    $d_1^V$ &          $d_2^V$ \\
    \end{bmatrix} = \\
    &= \renewcommand\arraystretch{1.5}\begin{bmatrix}[*2{C{80mm}}]
    $(a^Q a^K)_S \cdot a_1^V + (a^Q b^K)_S \cdot b_1^V + (a^Q d^K)_S \cdot d_1^V + 0$ & $(a^Q a^K)_S \cdot a_2^V + (a^Q b^K)_S \cdot b_2^V + (a^Q d^K)_S \cdot d_2^V + 0$ \\
    $(b^Q a^K)_S \cdot a_1^V + (b^Q b^K)_S \cdot b_1^V + (b^Q d^K)_S \cdot d_1^V + 0$ & $(b^Q a^K)_S \cdot a_2^V + (b^Q b^K)_S \cdot b_2^V + (b^Q d^K)_S \cdot d_2^V + 0$ \\
    $0$ & $0$ \\
    $(d^Q a^K)_S \cdot a_1^V + (d^Q b^K)_S \cdot b_1^V + (d^Q d^K)_S \cdot d_1^V + 0$ & $(d^Q a^K)_S \cdot a_2^V + (d^Q b^K)_S \cdot b_2^V + (d^Q d^K)_S \cdot d_2^V + 0$ \\
    \end{bmatrix}
\end{aligned}$}
\end{equation}
By comparing the final outputs, we can note that our method allows the contributions of the missing feature to be completely masked, making the most of the available data.

\newpage 

\section{Training of the models}\label{app:training}
 
For NAIM, we specified its architecture with an embedding dimension $d_e=6$, encoder layers $L=6$, and attention heads $h=3$~\cite{bib:transformer}. 
We configured the feed-forward layer size to 1000 neurons, and we eliminated the bias in the embedding of numerical and categorical features, to represent missing values as zero vectors.

During the training of DL models, since all the datasets work on a classification task, we opted to use the cross-entropy as loss function.
To enhance the training process, we employed a dynamic learning rate schedule that reduced the rate by a factor of $10$ whenever the loss stagnates for $25$ epochs. 
The training incorporated the Adam optimization algorithm with a consistent batch size of $32$ across all models and datasets.
For all the DL models we used a Glorot uniform initialization of the weights, which were trained all the models for 1500 epochs, using an early stopping rule with patience of $50$ epochs. 
\rr We set the initial 50 epochs as a warm-up period for parameter tuning for all models, except GRAPE for which $500$ warm-up epochs are assigned due to its lack of batch sampling. \bb 
Furthermore, we incorporated L1 and L2 regularization techniques to prevent overfitting.

All models and techniques under consideration are evaluated using their publicly available implementations. 
For XGBoost\footnote{\url{https://github.com/dmlc/xgboost/tree/master}}, GRAPE\footnote{\url{https://github.com/maxiaoba/GRAPE/tree/master}}, TabNet\footnote{\url{https://github.com/dreamquark-ai/tabnet}}, TabTransformer\footnotemark[3], and FTTransformer\footnotemark[3]\footnotetext[3]{\url{https://github.com/lucidrains/tab-transformer-pytorch}}, these implementations can be found on GitHub, whereas for the remaining models and imputation techniques, we used their implementations available in the Sklearn library\footnote[4]{\url{https://scikit-learn.org/stable}}. 
For all the competitors, we set their default parameters or those provided in the public implementations by authors because we are not interested in fine-tuning the models. 
Therefore, for the same reason, even in the NAIM implementation we did not perform parameter tuning, but we selected them according to the dimensionality of the data used.
Indeed, in \cite{bib:tune_param}, confirming the "No Free Lunch" theorem, the authors empirically observe that in many cases the use of tuned parameters cannot significantly outperform the default values of a classifier suggested in the literature. 

\newpage

\section{Results}\label{app:results}

\begin{figure}[!htbp]
\centering
\begin{adjustbox}{angle=90}
\begin{minipage}{.75\textheight}
\centering
\resizebox{\textwidth}{!}{
\begin{tabular}{ll|c|c|c|c|c|c|c|c|c|c|c|c|c|c|c|c|c|c}
\toprule
        & \multicolumn{1}{c}{} & \multicolumn{6}{c|}{\textbf{Train missing percentage:} 0\%} & \multicolumn{6}{c|}{\textbf{Train missing percentage:} 5\%} & \multicolumn{6}{c}{\textbf{Train missing percentage:} 10\%} \\\midrule[1pt]
        & & \multicolumn{6}{c|}{\textbf{Test missing percentage:}} & \multicolumn{6}{c|}{\textbf{Test missing percentage:}} & \multicolumn{6}{c}{\textbf{Test missing percentage:}} \\
\textbf{Model} & \textbf{Strategy} &                          0\%  &            5\%  &            10\% &            25\% &            50\% &            75\% &                         0\%  &            5\%  &            10\% &            25\% &            50\% &            75\% &                          0\%  &            5\%  &            10\% &            25\% &            50\% &            75\% \\
\midrule
NAIM & Intrinsic &               90.82 (3.67) &  \textbf{90.47 (3.69)} &  \textbf{89.98 (3.71)} &  \textbf{88.24 (3.78)} &  \textbf{84.25 (3.78)} &  \textbf{77.20 (3.43)} &               \textbf{90.82 (3.60)} &  \textbf{90.36 (3.64)} &  \textbf{89.85 (3.70)} &  \textbf{88.46 (3.62)} &  \textbf{83.66 (4.17)} &  \textbf{76.54 (4.32)} &                90.60 (3.71) &  \textbf{90.28 (3.67)} &  \textbf{89.80 (3.71)} &  \textbf{88.24 (3.77)} &  \textbf{83.79 (4.33)} &  76.40 (4.19) \\\midrule
\multirow{4}{*}{NAIM w/o regularization} & Intrinsic & 90.11 (3.97) & 88.01 (4.06) & 85.33 (4.46) & 79.55 (4.40) & 69.63 (3.90) & 61.57 (2.00) & 90.37 (3.72) & 89.41 (4.01) & 88.47 (4.28) & 85.32 (4.18) & 77.17 (4.57) & 67.23 (2.78) & 90.46 (3.52) & 89.48 (3.90) & 88.91 (3.76) & 86.07 (4.11) & 78.67 (4.89) & 68.70 (3.56) \\
& Constant & 88.57 (3.91) & 87.35 (3.72) & 85.65 (3.77) & 82.28 (3.00) & 74.83 (2.43) & 66.27 (2.21) & 88.73 (3.63) & 87.98 (3.62) & 87.21 (3.55) & 84.39 (3.43) & 78.13 (2.63) & 69.64 (1.85) & 88.65 (3.65) & 88.03 (3.54) & 87.24 (3.53) & 85.09 (3.27) & 79.37 (2.75) & 70.76 (1.58) \\
\bottomrule
\end{tabular}}

\vspace*{.5cm}

\resizebox{\textwidth}{!}{
\begin{tabular}{ll|c|c|c|c|c|c|c|c|c|c|c|c|c|c|c|c|c|c}
\toprule
        & \multicolumn{1}{c}{} & \multicolumn{6}{c|}{\textbf{Train missing percentage:} 25\%} & \multicolumn{6}{c|}{\textbf{Train missing percentage:} 50\%} & \multicolumn{6}{c}{\textbf{Train missing percentage:} 75\%} \\\midrule[1pt]
        & & \multicolumn{6}{c|}{\textbf{Test missing percentage:}} & \multicolumn{6}{c|}{\textbf{Test missing percentage:}} & \multicolumn{6}{c}{\textbf{Test missing percentage:}} \\
\textbf{Model} & \textbf{Strategy} &                          0\%  &            5\%  &            10\% &            25\% &            50\% &            75\% &                          0\%  &            5\%  &            10\% &            25\% &            50\% &            75\% &                          0\%  &            5\%  &            10\% &            25\% &            50\% &            75\% \\
\midrule
NAIM & Intrinsic &                \textbf{90.69 (3.43)} &  \textbf{90.31 (3.47)} &  \textbf{90.02 (3.36)} &  \textbf{88.90 (3.18)} &  83.99 (4.12) &  76.75 (3.97) &                89.66 (3.79) &  89.38 (3.76) &  89.13 (3.69) &  87.32 (4.14) &  83.17 (4.79) &  76.40 (4.34) &                88.32 (4.06) &  \textbf{87.89 (4.23)} &  87.65 (4.17) &  \textbf{85.94 (4.65)} &  83.20 (4.34) &  76.20 (5.04) \\\midrule
\multirow{4}{*}{NAIM w/o regularization} & Intrinsic & 89.85 (4.08) & 89.44 (4.06) & 88.68 (4.16) & 86.58 (4.20) & 80.88 (4.24) & 70.59 (3.61) & 89.35 (4.03) & 89.25 (3.75) & 88.51 (4.04) & 86.39 (4.34) & 81.69 (4.65) & 72.45 (4.48) & \textbf{88.61 (4.06)} & 87.77 (4.50) & \textbf{88.01 (3.90)} & 85.40 (4.97) & 81.62 (5.23) & 75.13 (4.66) \\
& Constant & 88.26 (3.70) & 87.92 (3.58) & 87.20 (3.61) & 85.18 (3.66) & 80.90 (3.32) & 72.30 (2.74) & 87.67 (3.74) & 87.24 (3.74) & 86.52 (3.80) & 84.69 (3.82) & 80.59 (3.84) & 74.47 (3.18) & 85.63 (3.70) & 85.16 (3.55) & 84.27 (3.74) & 82.42 (3.79) & 78.87 (4.23) & 72.26 (4.40) \\
\bottomrule
\end{tabular}}
\vspace{10pt}
\captionof{table}{Average AUC performance and standard error (in brackets) of the experiments across the $5$ different datasets. 
To facilitate the analysis we highlighted the best performance in each column in bold.}
\label{tab:TABELLONE_RISULTATI}
\end{minipage}
\end{adjustbox}
\end{figure}

\begin{figure}[!ht]
\centering
\begin{adjustbox}{angle=90}
\begin{minipage}{.75\textheight}
\centering
\resizebox{\textwidth}{!}{
\begin{tabular}{ll|c|c|c|c|c|c|c|c|c|c|c|c|c|c|c|c|c|c}
\toprule
        & \multicolumn{1}{c}{} & \multicolumn{6}{c|}{\textbf{Train missing percentage:} 0\%} & \multicolumn{6}{c|}{\textbf{Train missing percentage:} 5\%} & \multicolumn{6}{c}{\textbf{Train missing percentage:} 10\%} \\\midrule[1pt]
        & & \multicolumn{6}{c|}{\textbf{Test missing percentage:}} & \multicolumn{6}{c|}{\textbf{Test missing percentage:}} & \multicolumn{6}{c}{\textbf{Test missing percentage:}} \\
\textbf{Model} & \textbf{Strategy} &                          0\%  &            5\%  &            10\% &            25\% &            50\% &            75\% &                         0\%  &            5\%  &            10\% &            25\% &            50\% &            75\% &                          0\%  &            5\%  &            10\% &            25\% &            50\% &            75\% \\
\midrule
NAIM & Intrinsic &               91.60 (0.21) &  91.30 (0.20) &  90.98 (0.19) &  \textbf{89.58 (0.14)} &  \textbf{86.19 (0.17)} &  \textbf{79.02 (0.25)} &               91.61 (0.17) &  91.29 (0.14) &  90.93 (0.18) &  89.68 (0.17) &  \textbf{86.30 (0.23)} &  \textbf{79.19 (0.25)} &                91.50 (0.19) &  91.27 (0.20) &  90.97 (0.21) &  89.69 (0.21) &  86.05 (0.18) &  \textbf{79.18 (0.13)} \\\midrule
\multirow{4}{*}{NAIM w/o regularization} & Intrinsic & 91.46 (0.17) & 88.79 (0.42) & 85.69 (0.71) & 78.18 (1.38) & 68.92 (0.59) & 62.13 (1.21) & 91.36 (0.25) & 90.97 (0.28) & 90.47 (0.26) & 88.56 (0.33) & 83.62 (0.57) & 74.29 (0.80) & 91.32 (0.12) & 90.94 (0.08) & 90.40 (0.09) & 88.44 (0.20) & 83.65 (0.67) & 74.88 (0.95) \\
& Constant & 86.15 (0.30) & 85.29 (0.23) & 84.02 (0.25) & 80.70 (0.15) & 73.27 (0.30) & 64.34 (0.57) & 86.23 (0.26) & 85.60 (0.27) & 84.68 (0.24) & 81.66 (0.40) & 75.30 (0.42) & 65.85 (0.36) & 86.03 (0.24) & 85.28 (0.21) & 84.58 (0.14) & 81.74 (0.29) & 76.38 (0.14) & 67.38 (0.43)  \\
\bottomrule
\end{tabular}}

\vspace*{.5cm}

\resizebox{\textwidth}{!}{
\begin{tabular}{ll|c|c|c|c|c|c|c|c|c|c|c|c|c|c|c|c|c|c}
\toprule
        & \multicolumn{1}{c}{} & \multicolumn{6}{c|}{\textbf{Train missing percentage:} 25\%} & \multicolumn{6}{c|}{\textbf{Train missing percentage:} 50\%} & \multicolumn{6}{c}{\textbf{Train missing percentage:} 75\%} \\\midrule[1pt]
        & & \multicolumn{6}{c|}{\textbf{Test missing percentage:}} & \multicolumn{6}{c|}{\textbf{Test missing percentage:}} & \multicolumn{6}{c}{\textbf{Test missing percentage:}} \\
\textbf{Model} & \textbf{Strategy} &                          0\%  &            5\%  &            10\% &            25\% &            50\% &            75\% &                          0\%  &            5\%  &            10\% &            25\% &            50\% &            75\% &                          0\%  &            5\%  &            10\% &            25\% &            50\% &            75\% \\
\midrule
NAIM & Intrinsic &                91.44 (0.18) &  91.23 (0.17) &  90.91 (0.17) &  89.65 (0.14) &  86.27 (0.31) &  79.36 (0.29) &                91.00 (0.25) &  90.77 (0.27) &  90.41 (0.25) &  89.44 (0.26) &  86.27 (0.21) &  79.46 (0.24) &                90.50 (0.16) &  90.28 (0.17) &  89.97 (0.16) &  88.79 (0.23) &  85.88 (0.27) &  79.56 (0.25) \\\midrule
\multirow{4}{*}{NAIM w/o regularization} & Intrinsic & 91.31 (0.22) & 91.04 (0.23) & 90.68 (0.21) & 89.34 (0.27) & 85.65 (0.27) & 77.90 (0.36) & 90.95 (0.21) & 90.68 (0.22) & 90.23 (0.25) & 88.86 (0.29) & 85.02 (0.63) & 77.22 (0.92) & 90.49 (0.19) & 90.24 (0.20) & 89.84 (0.23) & 88.63 (0.23) & 85.53 (0.35) & 78.95 (0.39) \\
& Constant & 85.74 (0.30) & 85.25 (0.30) & 84.47 (0.31) & 82.01 (0.37) & 77.16 (0.42) & 68.59 (0.26) & 85.35 (0.24) & 84.78 (0.24) & 84.18 (0.28) & 81.78 (0.26) & 76.91 (0.24) & 69.05 (0.41) & 85.08 (0.23) & 84.57 (0.22) & 83.81 (0.25) & 81.65 (0.26) & 76.88 (0.26) & 69.14 (0.37) \\
\bottomrule
\end{tabular}}

\vspace{10pt} 
\captionof{table}{Average AUC performance and standard error (in brackets) of the experiments on the ADULT dataset.
To facilitate the analysis we highlighted the best performance in each column in bold.}
\label{tab:TAB_adult}
\end{minipage}
\end{adjustbox}
\end{figure}

\begin{figure}[!ht]
\centering
\begin{adjustbox}{angle=90}
\begin{minipage}{.75\textheight}
\centering
\resizebox{\textwidth}{!}{
\begin{tabular}{ll|c|c|c|c|c|c|c|c|c|c|c|c|c|c|c|c|c|c}
\toprule
        & \multicolumn{1}{c}{} & \multicolumn{6}{c|}{\textbf{Train missing percentage:} 0\%} & \multicolumn{6}{c|}{\textbf{Train missing percentage:} 5\%} & \multicolumn{6}{c}{\textbf{Train missing percentage:} 10\%} \\\midrule[1pt]
        & & \multicolumn{6}{c|}{\textbf{Test missing percentage:}} & \multicolumn{6}{c|}{\textbf{Test missing percentage:}} & \multicolumn{6}{c}{\textbf{Test missing percentage:}} \\
\textbf{Model} & \textbf{Strategy} &                          0\%  &            5\%  &            10\% &            25\% &            50\% &            75\% &                         0\%  &            5\%  &            10\% &            25\% &            50\% &            75\% &                          0\%  &            5\%  &            10\% &            25\% &            50\% &            75\% \\
\midrule
NAIM & Intrinsic &               94.32 (0.13) &  93.89 (0.14) &  \textbf{93.23 (0.17)} &  \textbf{91.35 (0.13)} &  \textbf{86.01 (0.24)} &  76.72 (0.14) &               94.27 (0.11) &  93.77 (0.15) &  93.21 (0.13) &  \textbf{91.25 (0.19)} &  85.72 (0.08) &  77.64 (0.42) &                94.00 (0.16) &  93.43 (0.11) &  92.95 (0.21) &  90.97 (0.25) &  86.03 (0.25) &  \textbf{77.55 (0.50)} \\\midrule
\multirow{4}{*}{NAIM w/o regularization} & Intrinsic & 93.56 (0.21) & 90.66 (0.77) & 87.73 (1.25) & 80.38 (2.37) & 70.07 (2.96) & 61.86 (2.44) & 93.62 (0.19) & 92.91 (0.17) & 92.12 (0.21) & 88.89 (0.36) & 80.61 (0.58) & 69.72 (0.68) & 93.53 (0.13) & 92.81 (0.17) & 91.83 (0.24) & 89.08 (0.35) & 82.51 (0.64) & 71.18 (0.78)\\
& Constant & 93.28 (0.21) & 92.55 (0.23) & 91.13 (0.39) & 87.96 (0.45) & 81.26 (0.51) & 71.58 (1.02) & 93.25 (0.19) & 92.35 (0.24) & 91.70 (0.19) & 88.75 (0.44) & 82.77 (0.60) & 73.90 (0.49) & 93.30 (0.22) & 92.53 (0.21) & 91.54 (0.33) & 89.13 (0.37) & 83.51 (0.31) & 73.99 (0.39) \\
\bottomrule
\end{tabular}}

\vspace*{.5cm}

\resizebox{\textwidth}{!}{
\begin{tabular}{ll|c|c|c|c|c|c|c|c|c|c|c|c|c|c|c|c|c|c}
\toprule
        & \multicolumn{1}{c}{} & \multicolumn{6}{c|}{\textbf{Train missing percentage:} 25\%} & \multicolumn{6}{c|}{\textbf{Train missing percentage:} 50\%} & \multicolumn{6}{c}{\textbf{Train missing percentage:} 75\%} \\\midrule[1pt]
        & & \multicolumn{6}{c|}{\textbf{Test missing percentage:}} & \multicolumn{6}{c|}{\textbf{Test missing percentage:}} & \multicolumn{6}{c}{\textbf{Test missing percentage:}} \\
\textbf{Model} & \textbf{Strategy} &                          0\%  &            5\%  &            10\% &            25\% &            50\% &            75\% &                          0\%  &            5\%  &            10\% &            25\% &            50\% &            75\% &                          0\%  &            5\%  &            10\% &            25\% &            50\% &            75\% \\
\midrule
NAIM & Intrinsic &                93.71 (0.13) &  93.27 (0.13) &  92.81 (0.18) &  91.15 (0.15) &  85.84 (0.30) &  77.45 (0.18) &                93.14 (0.11) &  92.67 (0.11) &  92.17 (0.17) &  90.24 (0.11) &  85.77 (0.30) &  77.00 (0.31) &                91.18 (0.24) &  90.86 (0.22) &  90.43 (0.14) &  88.97 (0.24) &  84.81 (0.23) &  78.20 (0.42) \\\midrule
\multirow{4}{*}{NAIM w/o regularization} & Intrinsic & 93.68 (0.12) & 93.12 (0.14) & 92.19 (0.19) & 89.47 (0.44) & 82.40 (0.75) & 71.56 (1.81) & 93.08 (0.19) & 92.52 (0.18) & 91.96 (0.25) & 89.19 (0.30) & 83.92 (0.47) & 74.24 (0.77) & 92.13 (0.26) & 91.63 (0.30) & 91.02 (0.28) & 89.16 (0.25) & 84.73 (0.35) & 76.43 (0.85) \\
& Constant & 92.70 (0.15) & 92.06 (0.21) & 90.98 (0.17) & 88.49 (0.13) & 82.68 (0.38) & 74.49 (0.49) & 92.29 (0.21) & 91.68 (0.26) & 90.77 (0.29) & 88.13 (0.26) & 82.72 (0.31) & 74.09 (0.38) & 91.42 (0.25) & 90.71 (0.25) & 89.95 (0.22) & 87.53 (0.26) & 82.97 (0.43) & 74.11 (0.76) \\
\bottomrule
\end{tabular}}

\vspace{10pt} 
\captionof{table}{Average AUC performance and standard error (in brackets) of the experiments on the BankMarketing dataset.
To facilitate the analysis we highlighted the best performance in each column in bold.}
\label{tab:TAB_bankmarketing}
\end{minipage}
\end{adjustbox}
\end{figure}

\begin{figure}[!ht]
\centering
\begin{adjustbox}{angle=90}
\begin{minipage}{.75\textheight}
\centering
\resizebox{\textwidth}{!}{
\begin{tabular}{ll|c|c|c|c|c|c|c|c|c|c|c|c|c|c|c|c|c|c}
\toprule
        & \multicolumn{1}{c}{} & \multicolumn{6}{c|}{\textbf{Train missing percentage:} 0\%} & \multicolumn{6}{c|}{\textbf{Train missing percentage:} 5\%} & \multicolumn{6}{c}{\textbf{Train missing percentage:} 10\%} \\\midrule[1pt]
        & & \multicolumn{6}{c|}{\textbf{Test missing percentage:}} & \multicolumn{6}{c|}{\textbf{Test missing percentage:}} & \multicolumn{6}{c}{\textbf{Test missing percentage:}} \\
\textbf{Model} & \textbf{Strategy} &                          0\%  &            5\%  &            10\% &            25\% &            50\% &            75\% &                         0\%  &            5\%  &            10\% &            25\% &            50\% &            75\% &                          0\%  &            5\%  &            10\% &            25\% &            50\% &            75\% \\
\midrule
NAIM & Intrinsic &               92.80 (0.23) &  \textbf{92.25 (0.27)} &  \textbf{91.46 (0.25)} &  \textbf{88.36 (0.40)} &  \textbf{82.23 (0.49)} &  \textbf{72.97 (0.95)} &               92.64 (0.24) &  \textbf{91.85 (0.22)} &  \textbf{90.99 (0.22)} &  \textbf{88.43 (0.06)} &  \textbf{82.01 (0.42)} &  \textbf{73.75 (0.53)} &                92.52 (0.26) &  \textbf{91.91 (0.30)} &  91.15 (0.30) &  \textbf{88.64 (0.36)} &  \textbf{82.88 (0.42)} &  72.57 (0.46) \\\midrule
\multirow{4}{*}{NAIM w/o regularization} & Intrinsic &               92.49 (0.24) &  90.79 (0.28) &  88.78 (0.38) &  83.42 (0.69) &  72.75 (1.39) &  63.15 (2.21) &               92.43 (0.15) &  91.38 (0.14) &  90.56 (0.18) &  86.43 (0.23) &  77.30 (0.73) &  64.94 (1.29) &                92.11 (0.34) &  91.14 (0.33) &  90.04 (0.35) &  86.97 (0.56) &  79.15 (0.37) &  67.77 (0.56) \\
\bottomrule
\end{tabular}}

\vspace*{.5cm}

\resizebox{\textwidth}{!}{
\begin{tabular}{ll|c|c|c|c|c|c|c|c|c|c|c|c|c|c|c|c|c|c}
\toprule
        & \multicolumn{1}{c}{} & \multicolumn{6}{c|}{\textbf{Train missing percentage:} 25\%} & \multicolumn{6}{c|}{\textbf{Train missing percentage:} 50\%} & \multicolumn{6}{c}{\textbf{Train missing percentage:} 75\%} \\\midrule[1pt]
        & & \multicolumn{6}{c|}{\textbf{Test missing percentage:}} & \multicolumn{6}{c|}{\textbf{Test missing percentage:}} & \multicolumn{6}{c}{\textbf{Test missing percentage:}} \\
\textbf{Model} & \textbf{Strategy} &                          0\%  &            5\%  &            10\% &            25\% &            50\% &            75\% &                          0\%  &            5\%  &            10\% &            25\% &            50\% &            75\% &                          0\%  &            5\%  &            10\% &            25\% &            50\% &            75\% \\
\midrule
NAIM & Intrinsic &                92.05 (0.32) &  91.40 (0.37) &  90.66 (0.28) &  \textbf{88.56 (0.49)} &  \textbf{83.16 (0.47)} &  73.15 (0.75) &                90.83 (0.47) &  90.36 (0.45) &  89.88 (0.53) &  87.74 (0.32) &  \textbf{82.92 (0.63)} &  74.07 (0.69) &                89.50 (0.74) &  89.08 (0.67) &  \textbf{88.65 (0.79)} &  \textbf{86.59 (0.62)} &  81.39 (0.70) &  73.62 (0.64) \\\midrule
\multirow{4}{*}{NAIM w/o regularization} & Intrinsic &                91.80 (0.39) &  91.08 (0.50) &  90.03 (0.52) &  87.15 (0.62) &  79.81 (0.66) &  68.51 (0.79) &                90.89 (0.34) &  90.14 (0.34) &  89.47 (0.38) &  87.05 (0.46) &  80.24 (0.34) &  70.49 (0.53) &                89.67 (0.55) &  88.98 (0.47) &  88.52 (0.54) &  85.99 (0.39) &  80.05 (0.64) &  70.99 (0.48) \\
& Constant & 89.56 (0.66) & 89.01 (0.63) & 88.17 (0.63) & 85.93 (0.50) & 80.93 (0.73) & 71.50 (0.58) & 89.01 (0.58) & 88.54 (0.51) & 87.68 (0.68) & 85.78 (0.82) & 80.34 (0.71) & 72.38 (0.90) & 80.78 (4.17) & 80.21 (4.13) & 79.70 (3.92) & 77.04 (3.72) & 73.24 (3.48) & 65.57 (2.32) \\
\bottomrule
\end{tabular}}

\vspace{10pt} 
\captionof{table}{Average AUC performance and standard error (in brackets) of the experiments on the OnlineShoppers dataset.
To facilitate the analysis we highlighted the best performance in each column in bold.}
\label{tab:TAB_onlineshoppers}
\end{minipage}
\end{adjustbox}
\end{figure}

\begin{figure}[!ht]
\centering
\begin{adjustbox}{angle=90}
\begin{minipage}{.75\textheight}
\centering
\resizebox{\textwidth}{!}{
\begin{tabular}{ll|c|c|c|c|c|c|c|c|c|c|c|c|c|c|c|c|c|c}
\toprule
        & \multicolumn{1}{c}{} & \multicolumn{6}{c|}{\textbf{Train missing percentage:} 0\%} & \multicolumn{6}{c|}{\textbf{Train missing percentage:} 5\%} & \multicolumn{6}{c}{\textbf{Train missing percentage:} 10\%} \\\midrule[1pt]
        & & \multicolumn{6}{c|}{\textbf{Test missing percentage:}} & \multicolumn{6}{c|}{\textbf{Test missing percentage:}} & \multicolumn{6}{c}{\textbf{Test missing percentage:}} \\
\textbf{Model} & \textbf{Strategy} &                          0\%  &            5\%  &            10\% &            25\% &            50\% &            75\% &                         0\%  &            5\%  &            10\% &            25\% &            50\% &            75\% &                          0\%  &            5\%  &            10\% &            25\% &            50\% &            75\% \\
\midrule
NAIM & Intrinsic &               \textbf{76.89 (1.19)} &  \textbf{76.53 (1.41)} &  \textbf{76.05 (1.28)} &  74.46 (1.70) &  71.71 (1.69) &  68.41 (1.90) &               \textbf{77.19 (0.91)} &  \textbf{76.63 (0.81)} &  76.00 (0.80) &  75.40 (1.22) &  69.31 (1.57) &  62.68 (2.01) &                76.56 (1.04) &  76.48 (0.96) &  75.89 (1.16) &  74.46 (1.36) &  68.62 (0.61) &  63.54 (2.41) \\\midrule
\multirow{4}{*}{NAIM w/o regularization} & Intrinsic & 74.91 (1.67) & 72.74 (2.10) & 68.80 (3.44) & 64.38 (1.09) & 56.16 (2.04) & 54.28 (2.73) & 76.23 (1.06) & 74.11 (1.37) & 72.06 (1.24) & 69.24 (1.22) & 59.64 (2.67) & 57.84 (2.11) & 77.19 (1.06) & 74.69 (1.48) & 74.75 (0.99) & 70.63 (2.01) & 59.94 (2.32) & 55.37 (0.72) \\
& Constant & 74.97 (1.18) & 74.38 (1.27) & 72.46 (1.25) & 72.54 (0.89) & 68.79 (1.14) & 60.90 (0.98) & 76.39 (1.67) & 75.70 (1.75) & 75.32 (1.35) & 73.07 (1.76) & 69.91 (2.59) & 65.89 (3.60) & 76.22 (1.18) & 76.16 (1.27) & 75.44 (1.52) & 74.88 (1.17) & 71.08 (1.07) & 68.01 (1.84) \\
\bottomrule
\end{tabular}}

\vspace*{.5cm}

\resizebox{\textwidth}{!}{
\begin{tabular}{ll|c|c|c|c|c|c|c|c|c|c|c|c|c|c|c|c|c|c}
\toprule
        & \multicolumn{1}{c}{} & \multicolumn{6}{c|}{\textbf{Train missing percentage:} 25\%} & \multicolumn{6}{c|}{\textbf{Train missing percentage:} 50\%} & \multicolumn{6}{c}{\textbf{Train missing percentage:} 75\%} \\\midrule[1pt]
        & & \multicolumn{6}{c|}{\textbf{Test missing percentage:}} & \multicolumn{6}{c|}{\textbf{Test missing percentage:}} & \multicolumn{6}{c}{\textbf{Test missing percentage:}} \\
\textbf{Model} & \textbf{Strategy} &                          0\%  &            5\%  &            10\% &            25\% &            50\% &            75\% &                          0\%  &            5\%  &            10\% &            25\% &            50\% &            75\% &                          0\%  &            5\%  &            10\% &            25\% &            50\% &            75\% \\
\midrule
NAIM & Intrinsic &                \textbf{77.86 (0.89)} &  77.39 (0.74) &  \textbf{77.65 (0.93)} &  \textbf{77.76 (1.72)} &  69.59 (0.92) &  64.77 (1.12) &                75.41 (0.80) &  75.31 (1.19) &  75.44 (1.38) &  72.07 (1.44) &  65.85 (1.21) &  62.30 (1.42) &                73.04 (1.15) &  71.97 (1.11) &  72.04 (1.52) &  68.67 (1.93) &  68.52 (0.69) &  59.30 (3.16) \\\midrule
\multirow{4}{*}{NAIM w/o regularization} & Intrinsic & 74.26 (1.41) & 74.02 (1.85) & 72.92 (0.97) & 70.87 (1.35) & 65.61 (1.07) & 57.92 (2.33) & 74.04 (1.24) & 75.23 (1.41) & 73.37 (0.81) & 70.30 (1.78) & 65.41 (1.18) & 56.71 (1.80) & 73.30 (1.69) & 70.69 (2.50) & 73.54 (1.59) & 66.78 (3.09) & 62.82 (2.57) & 60.50 (1.99) \\
& Constant & 75.69 (1.30) & 75.99 (1.24) & 75.33 (1.87) & 73.65 (2.15) & 71.81 (1.91) & 65.45 (1.91) & 74.80 (1.02) & 74.49 (1.19) & 73.60 (0.90) & 72.28 (1.06) & 69.74 (1.12) & 70.12 (1.31) & 75.02 (1.31) & 75.24 (1.40) & 73.33 (1.61) & 72.18 (1.44) & 68.40 (2.16) & 64.04 (1.92) \\
\bottomrule
\end{tabular}}

\vspace{10pt} 
\captionof{table}{Average AUC performance and standard error (in brackets) of the experiments on the SeismicBumps dataset.
To facilitate the analysis we highlighted the best performance in each column in bold.}
\label{tab:TAB_seismicbumps}
\end{minipage}
\end{adjustbox}
\end{figure}

\begin{figure}[!ht]
\centering
\begin{adjustbox}{angle=90}
\begin{minipage}{.75\textheight}
\centering
\resizebox{\textwidth}{!}{
\begin{tabular}{ll|c|c|c|c|c|c|c|c|c|c|c|c|c|c|c|c|c|c}
\toprule
        & \multicolumn{1}{c}{} & \multicolumn{6}{c|}{\textbf{Train missing percentage:} 0\%} & \multicolumn{6}{c|}{\textbf{Train missing percentage:} 5\%} & \multicolumn{6}{c}{\textbf{Train missing percentage:} 10\%} \\\midrule[1pt]
        & & \multicolumn{6}{c|}{\textbf{Test missing percentage:}} & \multicolumn{6}{c|}{\textbf{Test missing percentage:}} & \multicolumn{6}{c}{\textbf{Test missing percentage:}} \\
\textbf{Model} & \textbf{Strategy} &                          0\%  &            5\%  &            10\% &            25\% &            50\% &            75\% &                         0\%  &            5\%  &            10\% &            25\% &            50\% &            75\% &                          0\%  &            5\%  &            10\% &            25\% &            50\% &            75\% \\
\midrule
NAIM & Intrinsic &               98.50 (0.21) &   98.37 (0.17) &   98.18 (0.22) &   97.47 (0.17) &   95.13 (0.34) &  88.86 (0.54) &               98.40 (0.13) &  98.25 (0.14) &  98.14 (0.12) &  \textbf{97.53 (0.14)} &  94.96 (0.22) &  89.43 (0.40) &                98.44 (0.15) &   98.33 (0.18) &   98.06 (0.13) &   97.45 (0.10) &   95.37 (0.24) &  89.15 (0.18) \\\midrule
\multirow{4}{*}{NAIM w/o regularization} & Intrinsic & 98.13 (0.15) & 97.05 (0.20) & 95.65 (0.43) & 91.38 (1.04) & 80.25 (1.93) & 66.44 (0.92) & 98.20 (0.15) & 97.69 (0.09) & 97.15 (0.14) & 93.50 (0.50) & 84.69 (1.09) & 69.37 (0.76) & 98.14 (0.12) & 97.80 (0.12) & 97.19 (0.18) & 95.25 (0.28) & 88.08 (0.90) & 74.29 (0.77) \\
& Constant & 98.03 (0.22) & 96.09 (0.19) & 94.44 (0.39) & 89.25 (0.63) & 79.71 (1.11) & 71.44 (1.85) & 97.81 (0.15) & 97.24 (0.13) & 96.39 (0.21) & 93.30 (0.12) & 84.54 (0.97) & 74.13 (0.61) & 97.72 (0.18) & 97.04 (0.15) & 96.33 (0.29) & 94.10 (0.27) & 86.85 (0.72) & 75.04 (1.39) \\
\bottomrule
\end{tabular}}

\vspace*{.5cm}

\resizebox{\textwidth}{!}{
\begin{tabular}{ll|c|c|c|c|c|c|c|c|c|c|c|c|c|c|c|c|c|c}
\toprule
        & \multicolumn{1}{c}{} & \multicolumn{6}{c|}{\textbf{Train missing percentage:} 25\%} & \multicolumn{6}{c|}{\textbf{Train missing percentage:} 50\%} & \multicolumn{6}{c}{\textbf{Train missing percentage:} 75\%} \\\midrule[1pt]
        & & \multicolumn{6}{c|}{\textbf{Test missing percentage:}} & \multicolumn{6}{c|}{\textbf{Test missing percentage:}} & \multicolumn{6}{c}{\textbf{Test missing percentage:}} \\
\textbf{Model} & \textbf{Strategy} &                          0\%  &            5\%  &            10\% &            25\% &            50\% &            75\% &                          0\%  &            5\%  &            10\% &            25\% &            50\% &            75\% &                          0\%  &            5\%  &            10\% &            25\% &            50\% &            75\% \\
\midrule
NAIM & Intrinsic &                98.37 (0.12) &   98.24 (0.14) &   98.05 (0.08) &   97.36 (0.13) &   95.11 (0.44) &  89.03 (0.49) &                97.93 (0.22) &  97.77 (0.25) &  97.73 (0.27) &  97.13 (0.28) &  95.06 (0.21) &  89.16 (0.52) &                97.38 (0.24) &  97.25 (0.25) &  \textbf{97.14 (0.27)} &  \textbf{96.70 (0.29)} &  \textbf{95.38 (0.25)} &  90.34 (0.31) \\\midrule
\multirow{4}{*}{NAIM w/o regularization} & Intrinsic & 98.21 (0.12) & 97.95 (0.12) & 97.60 (0.15) & 96.08 (0.23) & 90.92 (0.63) & 77.07 (1.43) & 97.78 (0.17) & 97.70 (0.18) & 97.50 (0.13) & 96.54 (0.29) & 93.88 (0.49) & 83.60 (0.38) & 97.45 (0.16) & \textbf{97.31 (0.16)} & 97.11 (0.19) & 96.43 (0.23) & 94.54 (0.40) & 88.77 (0.56) \\
& Constant & 97.63 (0.13) & 97.30 (0.12) & 97.05 (0.15) & 95.81 (0.21) & 91.90 (0.42) & 81.48 (1.41) & 96.89 (0.33) & 96.69 (0.42) & 96.36 (0.50) & 95.49 (0.57) & 93.22 (0.51) & 86.71 (0.89) & 95.83 (0.42) & 95.08 (0.59) & 94.57 (0.56) & 93.69 (0.22) & 92.85 (0.18) & 88.45 (0.45) \\
\bottomrule
\end{tabular}}

\vspace{10pt} 
\captionof{table}{Average AUC performance and standard error (in brackets) of the experiments on the Spambase dataset.
To facilitate the analysis we highlighted the best performance in each column in bold.}
\label{tab:TAB_spambase}
\end{minipage}
\end{adjustbox}
\end{figure}

\clearpage

\section{Statistical analysis}\label{app:stats}

\tablename~\ref{tab:TABELLONE} presents the percentages of experiments in which NAIM is significantly better than the competitors. 
Additionally, the averages across rows and columns are provided at the right and bottom margins, respectively, to identify an average trend. 
The analysis of the row and column averages reveals that the win rate of NAIM significantly exceeds the loss rate, with a considerable minimum advantage of $7.8\%$ over the competitors. 
This observation, reinforced by the data in the bottom right-hand corner, which indicates that NAIM, on average, wins against its competitors in $58.7\%$ of the cases, while it only loses in $1.6\%$ of the cases, underlines NAIM's ability to provide competitive performance on tabular data without the need for any imputation strategy.

\begin{table*}[!ht]
    \centering
    \resizebox{\textwidth}{!}{
    \begin{tabular}{lc|cc|cc|cc|cc|cc||cc}
    \toprule
    & & \multicolumn{10}{c||}{\textbf{Datasets}} \\
     &      & \multicolumn{2}{c|}{ADULT} & \multicolumn{2}{c|}{BankMarketing} & \multicolumn{2}{c|}{OnlineShoppers} & \multicolumn{2}{c|}{SeismicBumps} & \multicolumn{2}{c||}{Spambase} & \multicolumn{2}{c}{Mean} \\
     \textbf{Model} &  \textbf{Strategy}  &  \% Win & \% Loss &         \% Win & \% Loss &           \% Win & \% Loss &        \% Win & \% Loss &    \% Win & \% Loss &  \% Win & \% Loss \\
\midrule[1pt]
\multirow{3}{*}{Adaboost} & Constant &   19.4 &   11.1 &          13.9 &    0.0 &            69.4 &    0.0 &          0.0 &    0.0 &     38.9 &    0.0 &   28.3 &    2.2 \\
     & KNN &   36.1 &    2.8 &          19.4 &    0.0 &           100.0 &    0.0 &          0.0 &    0.0 &     75.0 &    0.0 &   46.1 &    0.6 \\
     & MICE &   55.6 &    0.0 &           8.3 &    2.8 &            97.2 &    0.0 &         13.9 &    0.0 &     86.1 &    0.0 &   52.2 &    0.6 \\\midrule
\multirow{4}{*}{Decision Tree} & Constant &  100.0 &    0.0 &         100.0 &    0.0 &           100.0 &    0.0 &         97.2 &    0.0 &    100.0 &    0.0 &   99.4 &    0.0 \\
     & KNN &  100.0 &    0.0 &         100.0 &    0.0 &           100.0 &    0.0 &        100.0 &    0.0 &    100.0 &    0.0 &  100.0 &    0.0 \\
     & MICE &  100.0 &    0.0 &         100.0 &    0.0 &           100.0 &    0.0 &         97.2 &    0.0 &    100.0 &    0.0 &   99.4 &    0.0 \\
     & Intrinsic &  100.0 &    0.0 &         100.0 &    0.0 &            97.2 &    0.0 &        100.0 &    0.0 &    100.0 &    0.0 &   99.4 &    0.0 \\\midrule
\multirow{3}{*}{FTTransformer} & Constant &  100.0 &    0.0 &          19.4 &    0.0 &            52.8 &    0.0 &          0.0 &    0.0 &    100.0 &    0.0 &   54.4 &    0.0 \\
     & KNN &  100.0 &    0.0 &          38.9 &    0.0 &            94.4 &    0.0 &          0.0 &    0.0 &    100.0 &    0.0 &   66.7 &    0.0 \\
     & MICE &  100.0 &    0.0 &          19.4 &    0.0 &            80.6 &    0.0 &          0.0 &    0.0 &    100.0 &    0.0 &   60.0 &    0.0 \\\midrule
GRAPE & Intrinsic & 100.0 & 0.0 & 88.9 & 0.0 & 94.4 & 0.0 & 0.0 & 0.0 & 100.0 & 0.0 & 76.7 & 0.0 \\\midrule
\multirow{4}{*}{HistGradientBoost} & Constant &  100.0 &    0.0 &           2.8 &   19.4 &            88.9 &    0.0 &          5.6 &    0.0 &     19.4 &    0.0 &   43.3 &    3.9 \\
     & KNN &  100.0 &    0.0 &           8.3 &   13.9 &           100.0 &    0.0 &          0.0 &    0.0 &     36.1 &    2.8 &   48.9 &    3.3 \\
     & MICE &  100.0 &    0.0 &           8.3 &    8.3 &           100.0 &    0.0 &          0.0 &    0.0 &     38.9 &    0.0 &   49.4 &    1.7 \\
     & Intrinsic &    8.3 &   80.6 &           0.0 &   13.9 &            19.4 &    0.0 &          2.8 &    0.0 &      5.6 &    2.8 &    7.2 &   19.4 \\\midrule
\multirow{3}{*}{MLP} & Constant &  100.0 &    0.0 &         100.0 &    0.0 &           100.0 &    0.0 &         36.1 &    0.0 &     88.9 &    0.0 &   85.0 &    0.0 \\
     & KNN &  100.0 &    0.0 &         100.0 &    0.0 &           100.0 &    0.0 &         27.8 &    0.0 &     83.3 &    0.0 &   82.2 &    0.0 \\
     & MICE &  100.0 &    0.0 &         100.0 &    0.0 &           100.0 &    0.0 &         36.1 &    0.0 &     97.2 &    0.0 &   86.7 &    0.0 \\\midrule
\multirow{4}{*}{Random Forest} & Constant &  100.0 &    0.0 &          16.7 &    0.0 &            50.0 &    0.0 &          2.8 &    0.0 &     44.4 &    0.0 &   42.8 &    0.0 \\
     & KNN &  100.0 &    0.0 &          16.7 &    0.0 &            88.9 &    0.0 &          0.0 &    0.0 &     44.4 &    2.8 &   50.0 &    0.6 \\
     & MICE &  100.0 &    0.0 &          11.1 &    0.0 &            75.0 &    0.0 &          0.0 &    0.0 &     52.8 &    0.0 &   47.8 &    0.0 \\
     & Intrinsic &   66.7 &    0.0 &          19.4 &    0.0 &            13.9 &    0.0 &          2.8 &    0.0 &      2.8 &    2.8 &   21.1 &    0.6 \\\midrule
\multirow{3}{*}{SVM} & Constant &  100.0 &    0.0 &          72.2 &    0.0 &            88.9 &    0.0 &          0.0 &    0.0 &     77.8 &    0.0 &   67.8 &    0.0 \\
     & KNN &  100.0 &    0.0 &          52.8 &    0.0 &           100.0 &    0.0 &          0.0 &    0.0 &     86.1 &    0.0 &   67.8 &    0.0 \\
     & MICE &  100.0 &    0.0 &          63.9 &    0.0 &            88.9 &    0.0 &          0.0 &    0.0 &     72.2 &    0.0 &   65.0 &    0.0 \\\midrule
\multirow{3}{*}{TabNet} & Constant &  100.0 &    0.0 &          13.9 &    5.6 &            72.2 &    0.0 &          0.0 &    0.0 &    100.0 &    0.0 &   57.2 &    1.1 \\
     & KNN &  100.0 &    0.0 &          25.0 &    5.6 &           100.0 &    0.0 &          0.0 &    0.0 &     97.2 &    0.0 &   64.4 &    1.1 \\
     & MICE &  100.0 &    0.0 &          38.9 &    0.0 &            77.8 &    0.0 &          0.0 &    0.0 &    100.0 &    0.0 &   63.3 &    0.0 \\\midrule
\multirow{3}{*}{TabTransformer} & Constant &  100.0 &    0.0 &          13.9 &    0.0 &            30.6 &    0.0 &          0.0 &    0.0 &     94.4 &    0.0 &   47.8 &    0.0 \\
     & KNN &  100.0 &    0.0 &          13.9 &    0.0 &            80.6 &    0.0 &          0.0 &    0.0 &     83.3 &    0.0 &   55.6 &    0.0 \\
     & MICE &  100.0 &    0.0 &          38.9 &    0.0 &            80.6 &    0.0 &          0.0 &    0.0 &     88.9 &    0.0 &   61.7 &    0.0 \\\midrule
\multirow{4}{*}{XGBoost} & Constant &  100.0 &    0.0 &           0.0 &   11.1 &            22.2 &    0.0 &          0.0 &    0.0 &     41.7 &    0.0 &   32.8 &    2.2 \\
     & KNN &  100.0 &    0.0 &          16.7 &    5.6 &            88.9 &    0.0 &          0.0 &    0.0 &     38.9 &    0.0 &   48.9 &    1.1 \\
     & MICE &  100.0 &    0.0 &          22.2 &    8.3 &            83.3 &    0.0 &          0.0 &    0.0 &     63.9 &    0.0 &   53.9 &    1.7 \\
     & Intrinsic &   16.7 &   63.9 &          13.9 &    8.3 &            25.0 &    0.0 &         22.2 &    0.0 &     33.3 &    0.0 &   22.2 &   14.4 \\\midrule\midrule
\multicolumn{2}{c|}{Mean} &   88.7 & 4.5 & 39.4 & 2.9 & 78.9 & 0.0 & 15.6 & 0.0 & 71.2 & 0.3 & 58.7 & 1.6 \\
\bottomrule
\end{tabular}}
\caption{Percentages of wins and losses of NAIM correct predictions compared with the competitors using the Wilcoxon signed-rank test.}\label{tab:TABELLONE}
\end{table*} 


\end{document}